\documentclass{article}

\usepackage[utf8]{inputenc} %
\usepackage[T1]{fontenc}    %
\usepackage{microtype}
\usepackage{graphicx}
\usepackage{booktabs} %
\usepackage[aboveskip=2pt]{subcaption} %

\usepackage{hyperref}

\usepackage[accepted]{icml2023}

\usepackage[utf8]{inputenc} %
\usepackage[T1]{fontenc}    %
\usepackage{url}            %
\usepackage{booktabs}       %
\usepackage{amsfonts,amsmath}       %
\usepackage{nicefrac}       %
\usepackage{microtype}      %
\usepackage{xcolor}         %
\usepackage{graphicx}

\definecolor{mycolor0}{rgb}{0.2667,0.4471,0.7098}
\definecolor{mycolor1}{rgb}{0.1647,0.6706,0.3804}
\definecolor{mycolor2}{rgb}{0.8275,0.2627,0.3059}
\definecolor{mycolor3}{rgb}{0.5216,0.4392,0.7176}
\definecolor{mycolor4}{rgb}{0.8118,0.7255,0.4118}
\definecolor{mycolor5}{rgb}{0.2745,0.7176,0.8157}

\usepackage{xspace}
\newcommand*{\eg}{{\it e.g.}\@\xspace}
\newcommand*{\ie}{{\it i.e.}\@\xspace}

\newcommand*{\etc}{{\it etc.}\@\xspace}

\usepackage{tikz,pgfplots}
\usetikzlibrary{shapes,shapes.arrows,positioning,spy,intersections,shapes.geometric,calc}
\usepgfplotslibrary{statistics}

\usepackage[capitalize,nameinlink]{cleveref}
\crefname{section}{Sec.}{Secs.}
\crefname{proposition}{Prop.}{Props.}
\crefname{lemma}{Lem.}{Lems.}
\crefname{model}{Mod.}{Mods.}
\crefname{appendix}{App.}{Apps.}
\crefname{algorithm}{Alg.}{Algs.}

\usepackage{booktabs} %
\usepackage{subcaption} %

\renewcommand{\paragraph}[1]{{\bf #1}~~}

\newlength\figureheight
\newlength\figurewidth
\newlength\subfigurewidth

\usepackage{bm}
\def\vx{\bm{x}}
\def\vz{\bm{z}}
\def\X{\mathcal{X}}
\def\MX{\bm{X}}
\def\MZ{\bm{Z}}

\def\R{\mathbb{R}}
\def\Y{\mathcal{Y}}
\def\vy{\bm{y}}
\DeclareMathOperator{\PM}{PM}
\DeclareMathOperator{\RM}{RM}
\DeclareMathOperator{\Ex}{\mathbb{E}}
\DeclareMathOperator{\Var}{\mathbb{V}ar}

\pgfplotsset{compat=1.16}

\begin{document}

\twocolumn[

\icmltitle{Disentangling Model Multiplicity in Deep Learning}

\icmlsetsymbol{equal}{*}

\begin{icmlauthorlist}
\icmlauthor{Ari Heljakka}{comp}
\icmlauthor{Martin Trapp}{yyy}
\icmlauthor{Juho Kannala}{yyy}
\icmlauthor{Arno Solin}{yyy}
\end{icmlauthorlist}

\icmlaffiliation{yyy}{Department of Computer Science, Aalto University, Espoo, Finland}
\icmlaffiliation{comp}{GenMind Ltd, Helsinki, Finland (Contributed all theory. Work done while in Aalto University.)~}

\icmlcorrespondingauthor{Ari Heljakka}{heljakka@iki.fi}

\icmlkeywords{Machine Learning, ICML}

\vskip 0.3in
]

\printAffiliationsAndNotice{} %

\begin{abstract}

Model multiplicity is a well-known but poorly understood phenomenon that undermines the generalisation guarantees of machine learning models. It appears when two models with similar training-time performance differ in their predictions and real-world performance characteristics. This observed `predictive' multiplicity (PM) also implies elusive differences in the internals of the models, their `representational' multiplicity (RM). We introduce a conceptual and experimental setup for analysing RM by measuring activation similarity via singular vector canonical correlation analysis (SVCCA). We show that certain differences in training methods systematically result in larger RM than others and evaluate RM and PM over a finite sample as predictors for generalizability. We further correlate RM with PM measured by the variance in i.i.d.\ and out-of-distribution test predictions in four standard image data sets. Finally, instead of attempting to eliminate RM, we call for its systematic measurement and maximal exposure. 
\looseness-1
\end{abstract}

\section{Introduction}
Machine learning (ML) models are typically underdetermined by data. This fact is often poorly understood, imprecisely conceptualized, and only superficially measured. Consequently, one often mistakes the observed success of a model for proof that it has somehow ‘captured’ the relevant features of its target, and that one has converged upon the single ‘right’ model. Due to this oversight, surprising practical problems with real-world model generalization may appear.
As ML models are increasingly deployed into real-world environments, it has become common to find that a model that worked well in the tests faces various failures in real-world scenarios. The primary textbook explanation for this disconnect is that the training data of the model and the deployment-time data were not generated from the same distribution (`non-i.i.d.'), rendering the predictions unreliable.\looseness-1

Our focus is on the more elusive yet prevalent case in which model variants with similar training-time performance behave very \emph{differently} from each other on individual held-out test data samples. Here, the data available at training time is insufficient to justify sound model selection among the model variants. This phenomenon has been approached with concepts such as non-identifiability, underspecification~\citep{damour2020underspecification}, Rashomon set \citep{breiman2001statistical,fisher2019models}, arbitrariness-1 \citep{heljakka0}, and predictive multiplicity~\citep{marx2020predictive}. Even though the {\em model selection} problem appears well-grounded in data analysis \citep{modelsel}, the established techniques assume a well-defined likelihood measure, which in the case of neural networks often does not exist.
In short, the problem occurs when risk-equivalent model variants have different internal representations of the same data. We call these %
differences {\em representational multiplicity} (RM) and propose to measure RM in terms of %
correlations of activations across model variants. These differences, in turn, may bring about the observed spread of predictions across the variants of the same model, the definition of predictive multiplicity \citep[PM,][]{marx2020predictive}.
The multiplicity may result from
seemingly random factors, 
which limits the reliability of inferences for explanatory purposes and undermines the very notion of a `best model for the data'.

\begin{figure*}[t]
\centering\large

  \tikzstyle{mystar} = [fill=mycolor0,draw=mycolor0, star, star points=5, star point ratio=2.25, inner sep=3pt,rounded corners=1pt,anchor=center]
  \tikzstyle{mysquare} = [fill=mycolor3,draw=mycolor3,inner sep=9pt,rounded corners=0pt,anchor=center]
  \tikzstyle{mycircle} = [fill=mycolor1!50,draw=mycolor1,circle,inner sep=5pt,anchor=center]
  \tikzstyle{mytriangle} = [fill=mycolor2, isosceles triangle, isosceles triangle apex angle=60,rotate=-30,inner sep=4.25pt,rounded corners=2pt,anchor=center,shape border uses incircle]
  \tikzstyle{myarrow} = [draw=black!20, single arrow, minimum height=7mm, minimum width=2mm, single arrow head extend=2mm, fill=black!20, anchor=center, rotate=0, inner sep=3pt, rounded corners=1pt]  
  \tikzstyle{myfoo} = [fill=mycolor0, star, star points=4, star point ratio=2.25, inner sep=3pt,rounded corners=1pt,anchor=center]  
  
\begin{subfigure}[b]{0.68\textwidth}
\centering
\scalebox{.55}{%
\begin{tikzpicture}[inner sep=0,outer sep=0pt]

\newcommand{\modelblock}[6]{
    \draw[fill=black!10,draw=black!20,rounded corners=1mm, line width=3pt] ($(#1) + (2,-1.5)$) rectangle ++(3,3);
    \node[align=left] at ($(#1) + (2.6,-1.3)$) {\small #6};
    \node[#2] at ($(#1) + (2.75,0.75)$) {};
    \node[#3] at ($(#1) + (3.5,0)$) {}; 
    \node[#4] at ($(#1) + (4.25,-0.75)$) {}; 
    \foreach \x [count=\i from 0] in {#5}
      \node at ($(#1) + (2.75+0.75*\i,0.75-0.75*\i)$) {\bfseries \footnotesize\color{white} \x};
}

\newcommand{\outputblock}[3]{
    \node[] at ($(#1) + (7,2)$) {#3};
    \node[myarrow] (arrow) at ($(#1) + (6,0)$) {};   
    \foreach \x [count=\i] in {#2}
        \node at ($(#1) + (7,2-\i)$) {{\bfseries \x}};
}

\newcommand{\inputblock}[2]{
    \node[] at ($(#1) + (0,2)$) {Input};
    \node[mysquare] at ($(#1) + (0,1)$) {};
    \node[mycircle] at ($(#1) + (0,1)$) {};    
    \node[mytriangle] at ($(#1) + (0,0)$) {};
    \node[#2] at ($(#1) + (0,-1)$) {};
    \node[myarrow] (arrow1) at ($(#1) + (1,0)$) {};
    \node at ($(#1) + (-.6,1)$) {{\bfseries A}};
    \node at ($(#1) + (-.6,0)$) {{\bfseries B}};
    \node at ($(#1) + (-.6,-1)$) {{\bfseries C}};
}

  \newcommand{\block}[8]{%
    \node[] at ($(#2) + (0,2)$) {Input};
    \node[mysquare] (#1-a) at ($(#2) + (0,1)$) {};
    \node[mycircle] at ($(#2) + (0,1)$) {};    
    \node[mytriangle] (#1-b) at ($(#2) + (0,0)$) {};
    \node[#8] (#1-c) at ($(#2) + (0,-1)$) {};
    \node[myarrow] (#1-arrow1) at ($(#2) + (1,0)$) {};
    \node at ($(#2) + (-.6,1)$) {{\bfseries A}};
    \node at ($(#2) + (-.6,0)$) {{\bfseries B}};
    \node at ($(#2) + (-.6,-1)$) {{\bfseries C}};
    \draw[fill=black!10,draw=black!20,rounded corners=1mm, line width=3pt] ($(#2) + (2,-1.5)$) rectangle ++(3,3);
    \node[myarrow] (#1-arrow2) at ($(#2) + (6,0)$) {};    
    \node[#3] at ($(#2) + (2.75,0.75)$) {};
    \node[#4] at ($(#2) + (3.5,0)$) {}; 
    \node[#5] at ($(#2) + (4.25,-0.75)$) {}; 
    \foreach \x [count=\i from 0] in {#6}
      \node at ($(#2) + (2.75+0.75*\i,0.75-0.75*\i)$) {\bfseries \footnotesize\color{white} \x};
    \node[] at ($(#2) + (7,2)$) {Output};
    \foreach \x [count=\i] in {#7}
      \node at ($(#2) + (7,2-\i)$) {{\bfseries \x}};
  }
  
  \inputblock{0,2.75}{mystar}
  \modelblock{0,1}{mycircle}{mytriangle}{mystar}{A,B,D}{Model 2}
  \outputblock{0,1}{A,B,D}{~}

  \modelblock{0,4.5}{mysquare}{mytriangle}{mystar}{A,B,D}{Model 1}
  \outputblock{0,4.5}{A,B,D}{Output}

  \inputblock{9.5,2.75}{mystar}
  \modelblock{9.5,1}{mycircle}{mytriangle}{mystar}{A,C,C}{Model 2}
  \outputblock{9.5,1}{A,C,C}{~}

  \modelblock{9.5,4.5}{mysquare}{mytriangle}{mystar}{A,B,B}{Model 1}
  \outputblock{9.5,4.5}{A,B,B}{Output}

  \draw[line width=6pt,draw=black!20,rounded corners=1pt] (8,6) -- (8,-7);
  \draw[line width=6pt,draw=black!20,rounded corners=1pt] (-1,-2) -- (17,-2);
  \node[fill=white,inner sep=6mm] at (8,-2) {\Large\bf Accuracy $=\nicefrac{\bf2}{\bf3}$};

  \block{E}{0,-5}{mysquare}{mytriangle}{}{A,B,~}{A,B,?}{mystar}
  \block{F}{9.5,-5}{mysquare}{mytriangle}{mystar}{A,B,C}{A,B,?}{myfoo}

  \node at (3.5,-1.25) {High RM, Low PM};
  \node at (13,-1.25) {High RM, High PM};
  \node at (3.5,-7.25) {Epistemic uncertainty};
  \node at (13,-7.25) {Aleatoric uncertainty};
  
\end{tikzpicture}}\\[5pt]
\caption{Similar accuracy admits many forms of multiplicity and uncertainty 
}
\label{fig:teaser-a}
\end{subfigure}
\hfill
\begin{subfigure}[b]{0.31\textwidth}
\centering
\scalebox{.55}{%
\begin{tikzpicture}[inner sep=0,outer sep=0pt]

  \newcommand{\block}[5]{
    \draw[fill=black!10,draw=black!20,rounded corners=1mm, line width=3pt] (#1) rectangle ++(3.5,1.5);
    \node[#2,scale=1.5] at ($(#1) + (1,.75)$) {};
    \node[#3,scale=1.5] at ($(#1) + (2.5,.65)$) {};
    \node[#4,scale=1.5] at ($(#1) + (1,.75)$) {};    
    \coordinate (#5) at ($(#1) + (4,.75)$);
  }

  \block{0,0}{mysquare}{mytriangle}{}{1}
  \block{0,-2}{mysquare}{mytriangle}{}{2}
  \block{0,-4}{mysquare}{mytriangle}{}{3}
  \block{0,-7}{mysquare}{mytriangle}{}{4}
  \block{0,-9}{mycircle}{mytriangle}{}{5}
  \block{0,-11}{mysquare}{mytriangle}{mycircle}{6}

  \node[right= 3mm of 1] {\LARGE Model 1};
  \node[right= 3mm of 2] {\LARGE Model 2};  
  \node[right= 3mm of 3] {\LARGE Model 3};  
  \node[right= 3mm of 4] {\LARGE Model 1};
  \node[right= 3mm of 5] {\LARGE Model 2};  
  \node[right= 3mm of 6] {\LARGE Model 3};  
  \node[rotate=90, align=center] at (-1,-1.25) {\Large Training strategy 1 \\ Lower characteristic RM};
  \node[rotate=90, align=center] at (-1,-8.25) {\Large Training strategy 2 \\ Higher characteristic RM};  
  
\end{tikzpicture}}\\[-2pt]~
\caption{ Training strategy affects RM
}
\label{fig:teaser-b}
\end{subfigure}
\caption{(a)~Even under \emph{identical training} setup, and controlled for the same empirical risk, model variants may represent the input data differently (high RM) due to different dependencies on learned features (\protect\tikz[baseline=-.5ex]\protect\node[mystar,scale=.5]{};, \protect\tikz[baseline=-.5ex]\protect\node[mysquare,scale=.5]{};, \protect\tikz[baseline=-.5ex]\protect\node[mycircle,scale=.5]{};, \protect\tikz[baseline=-.5ex]\protect\node[mytriangle,scale=.5]{};)  when classifying the three input samples with true class labels A, B, C and the true features indicated under `Input' column. The 3rd input has true feature \protect\tikz[baseline=-.5ex]\protect\node[mystar,scale=.5]{}; 
 with true class C \etc These differences may or may not be observable at the outputs (low or high PM, respectively). (b)~Different training strategies may systematically yield different degrees of RM. Higher RM indicates correspondingly higher {\em hidden risk} for generalization performance.}
\label{fig:teaser}
\vspace*{-3pt}
\end{figure*}
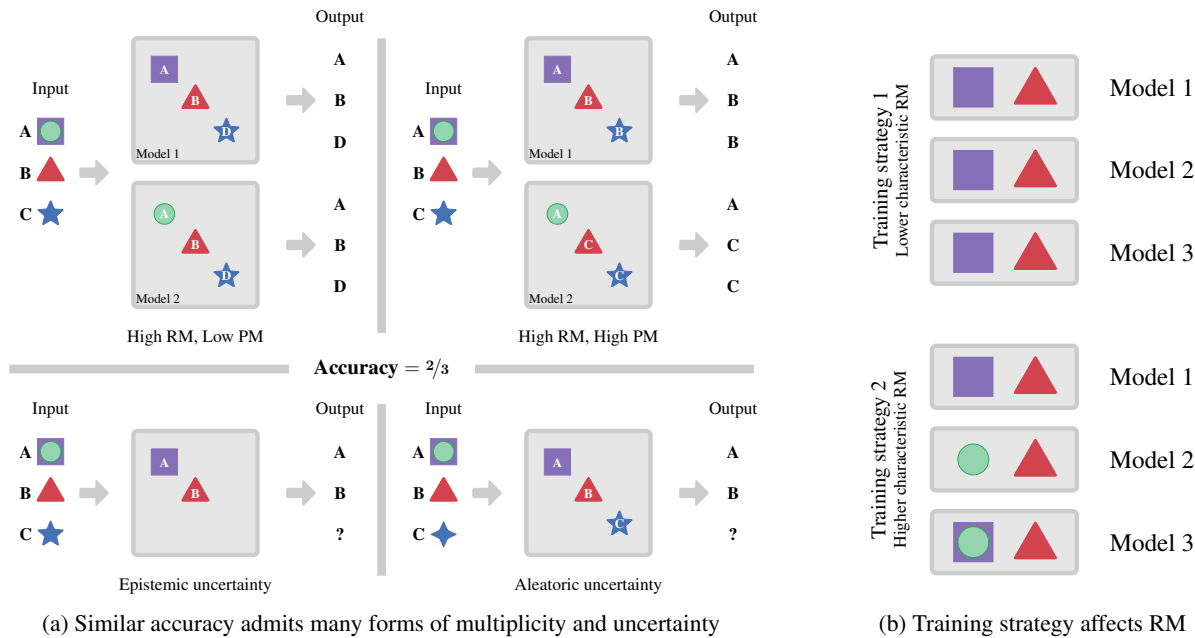

For deterministic pathways within a network, variations in predictive outputs imply variance in the intermediate representations. Hence, the presence of PM implies the presence of RM. The opposite is not the case since, for a given sample, two different representations can still lead to the same prediction. PM only measures the observable predictive differences \emph{as a function of the test samples at hand}, constrained by RM. %
Even in the absence of observed PM, any presence of RM implies that there exist other potential inputs that would also yield observable PM. This pivotal distinction fails to align with traditional notions of uncertainty defined in terms of observables (see \cref{fig:teaser-a}).

Many prior works approach model multiplicity as a nuisance, calling for its {\em elimination} or {\em reduction to uncertainty}. In contrast, we approach each internal representation as an alternative `compression' of the data, encapsulating hitherto unleveraged additional information. RM can be considered a {\em hidden risk} for future generalization performance. As such, for model variants with equivalent {\em observed} empirical risk, RM should be maximally {\em exposed} rather than minimized.

We look for possible systematic relationships between certain training strategies and the magnitude of RM \citep[suggested also in][]{raghu2017svcca,procca}.
Our vision data set results confirm a strong correlation between RM and learning rate / batch size.
Furthermore, irrespective of the well-known association between larger batch sizes and worse generalization via sharper minima \citep{hochreiter,keskar,trainlong}, larger batch sizes appear strongly correlated with lower PM.
Perhaps due to the combination of extra computational costs and somewhat muddled prior conceptual treatment, these empirically observable differences are often ignored
(see \cref{fig:teaser-b}).
Our emphasis on the irreducibility of RM aims to challenge the discourse based on a one-dimensional notion of `the best' model for the data and equivalent-sounding notions like `the predictor that encodes the \emph{right} structure' \citep{damour2020underspecification}, `[models that contain] the \emph{true} data generating process' \citep{fisher2019models} or even `true data generating model' \citep{modelsel}. This terminology falsely suggests one can cross and dissolve the categorical gap between the model and the modelled, leading to a single `correct' representation. However, in general, representational multiplicity can only be hidden, not eliminated. %

We summarize the contributions of this paper as follows.
{\em (i)}~We present a well-defined conceptual and experimental setup for analyzing the critical phenomenon of disentangled representational and predictive multiplicity (RM - PM). %
{\em (ii)}~We show the significance of observing both RM and PM through experiments and find various regularities between hyper-parameter values and RM. We design an experiment to empirically show that RM cannot be reduced to PM. 
{\em (iii)}~We introduce the \emph{confabulation matrix} as a straight-forward tool for visualizing multiplicity.

\section{Related Work}
\label{related}
The phenomenon of representative multiplicity in models has been studied as, for example, the underdetermination of theories in the philosophy of science and epistemology \citep[\eg,][]{duhem1954aim,newton1980underdetermination,laudan1991empirical}, and system non-identifiability in statistics and economics \citep[\eg,][]{dawid1979conditional,rothenberg1971identification,gelfand1999identifiability}. The implications for modern (very large) computational models are less obvious.
From a modern standpoint, the seminal paper by \citet{breiman2001statistical} %
contrasts two conceptions of models. First, the models that are structurally designed and selected for explanatory purposes, and second, the models optimized for prediction. In the latter case, the data admit a set of equally valid model parametrizations---the \textit{Rashomon set}--- and alternative competing explanations.

In computer science, %
the problem has been directly approached in the context of loss landscape analysis \citep[\eg,][]{garipov2018loss, izmailov2018averaging, chaudhari2019entropy, nakkiran2021deep}, overparametrization \citep[\eg,][]{belkin2019reconciling}, model averaging \citep[\eg,][]{izmailov2018averaging, wilson2020bayesian}, domain adaption and interpretability, and ensembling \citep{lakshminarayanan2017simple,fort2019deep}. Risk-equivalent modes have also been examined from the point of view of accuracy-diversity tradeoffs \citep{fort2019deep} as well as accuracy-reproducibility trade-offs \citep{shamir2020a}.
During recent years, empirical consequences of multiplicity have been examined in pathological cases such as spurious correlations and shortcut learning \citep{scholkopf2022causality, arjovsky2019invariant}. \citet{damour2020underspecification} examine the prevalence and consequences of multiplicity in practical ML pipelines that admit `underspecification'. %
Underspecification and Rashomon sets have directly been leveraged in \citet{semenova2019} with the goal of selecting models within the set based on criteria such as sparsity or monotonicity along an important set of features. In a similar vein, multiplicity can also be leveraged to estimate variable importance \citep{fisher2019models}.

The alternative notions of `underspecification', `identifiability', and Rashomon sets are closely related to the same scenario. To capture the internal and external aspects of the problem, the notion of representational multiplicity (RM) was introduced in this work to supplement \emph{predictive} multiplicity (PM) introduced in \citet{marx2020predictive}, a distinction often obscured in prior works. Further, our PM definition accounts for the actual variance of predictions, whereas, \eg, \citet{damour2020underspecification} only measure multiplicity as the \textit{variation in total accuracy}, ignoring the distribution in the \textit{kinds} of errors the model variants make.

For large models, `underspecification' as a concept problematically seems to imply that a `full' specification is possible. In \citet{damour2020underspecification}, `credible intuitive bias' is effectively defined as a way to eliminate the underspecification. This concept may be valuable in terms of segmenting the ML workflow, but it appears circular: we `solve' underspecification -- originally defined in terms of unobserved downstream environment -- by improving our pipeline so as to deal with the \textit{observed} problems. Shifting previously unobserved issues into the realm of observed issues surely fails to exhaust the realm of remaining unobserved issues.

Prior works often intrinsically posit the existence of a unique correct solution model, rendering any observed multiplicity into nuisance or noise to be minimized. This position can be defended if one can plausibly unearth the underlying causal properties via designing a set of interventions to that end. See \citet{damour2020underspecification} for a large-scale empirical evaluation and \citet{romeijn2018} for theoretical analysis. Yet, this rarely happens except in the very theory of causal inference itself \citep[see, \eg,][]{scholkopf2022causality}. In Bayesian and ensemble analyses \citep[\eg,][]{barber1998ensemble,fort2019deep,izmailov2018averaging}, the variance is characteristically treated in aggregate, usually in parameter space. Finally, as a matter of standard ML practice, the variance is reduced to a scalar for generalization error, even in the in-depth analysis of the topic in \citet{raghu2017svcca}.

Principled approaches to model selection exist in Bayesian statistics, with methods such as BIC and AIC routinely used to compare models. However, in practice, they are not applicable to deep neural networks \citep{modelsel} that perform equally well on the test data set. The same holds for techniques such as majority voting or randomization procedures \citep{germain2016pac}.
Prior works that focus on \textit{mitigating} the multiplicity in practice \citep{damour2020underspecification, arjovsky2019invariant, scholkopf2022causality} are valuable as long as we can assume that the information is available to resolve the multiplicity. Our work considers the general scenario in which we have no way of knowing whether this is the case nor of distinguishing one model from another, which renders all RM variations informative. Following the same line of reasoning, while it may be useful to marginalize over all model variants in parameter space \citep{wilson2020bayesian}, that may risk throwing away relevant information.

To avoid this, the quantification of RM in the network activation space is critical. %
We chose SVCCA \citep{raghu2017svcca} to efficiently compare the similarity of distributed representations across networks. SVCCA was shown to be capable of comparing different architectures and surfacing detailed learning dynamics, such as the order in which the layers `solidify' during learning. In comparison to regular canonical correlation analysis (CCA), SVCCA provides the importance of each direction in the original activation space. \citet{hinton2018} improve the reliability of the method when the number of evaluation data points is small.

The treatment of predictive multiplicity (PM) in \citet{marx2020predictive} is closest in spirit to our work (but also see the `prediction variation' metric in \citet{chen2020} and related \citet{dusenberry2020}). They call for reporting PM in quantitative measurements in the same way as the canonical test error. However, while they build upon `the Rashomon effect' (effectively equivalent to RM), they focus on the practical consequences for predictions and do not quantitatively address the PM-RM distinction. Our work provides a more complete yet concise picture that draws from the methodology of \citet{damour2020underspecification}, the explicit metrics for RM from \citet{raghu2017svcca} and the conceptual import for PM from \citet{marx2020predictive}. Most importantly, we hope to motivate incorporating these conceptual distinctions in regular ML practises.\looseness-1

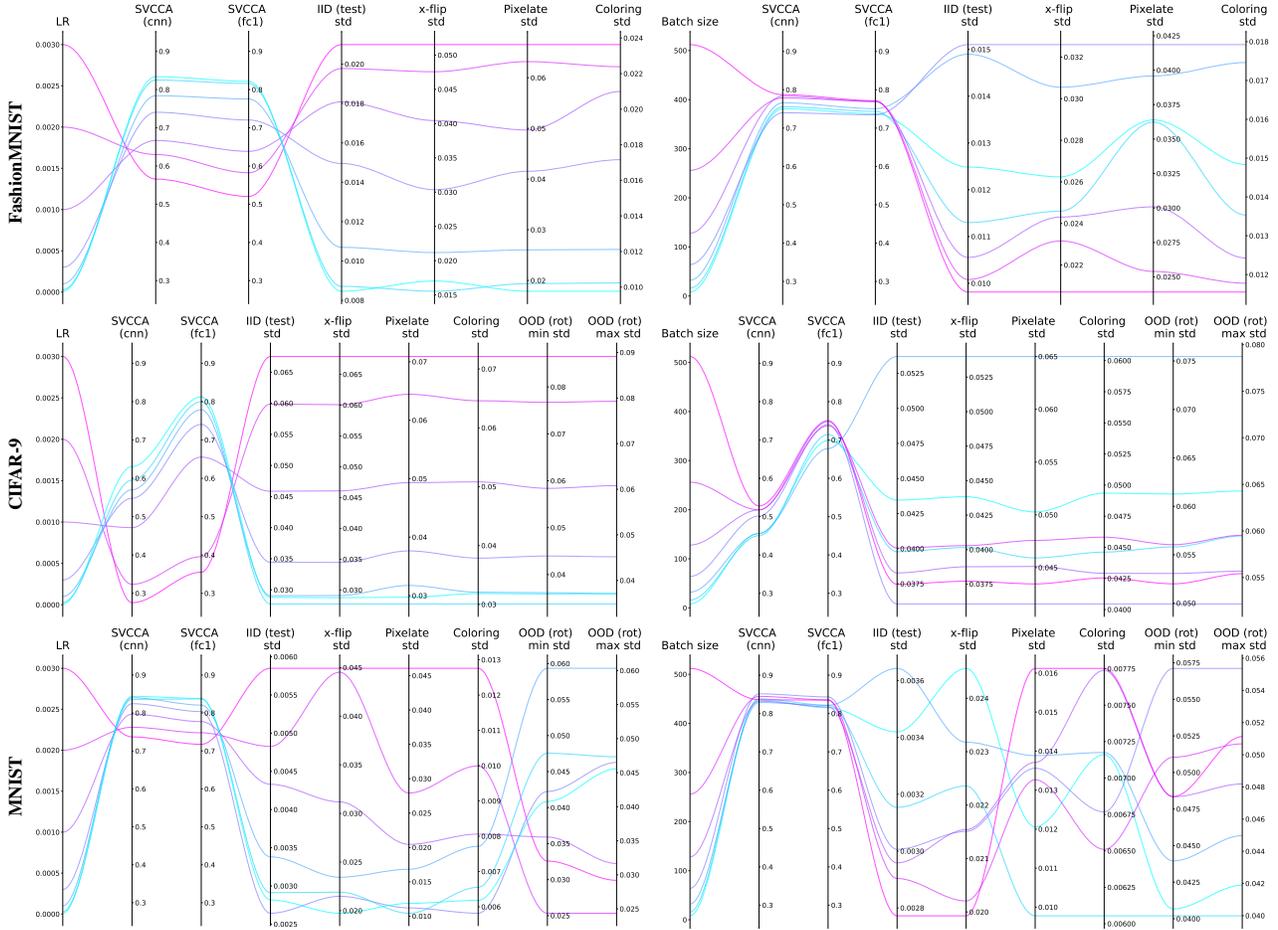
\begin{figure*}[t!]
\centering
\setlength{\figurewidth}{.23\textwidth}
\begin{tikzpicture}[inner sep=0]
    \node[rotate=90,minimum width=.99\figurewidth,inner sep=1pt,rounded corners=1pt] at (\figurewidth,.55\figurewidth) {\bf \scriptsize FashionMNIST};
\end{tikzpicture}
\includegraphics[width=0.48\textwidth]{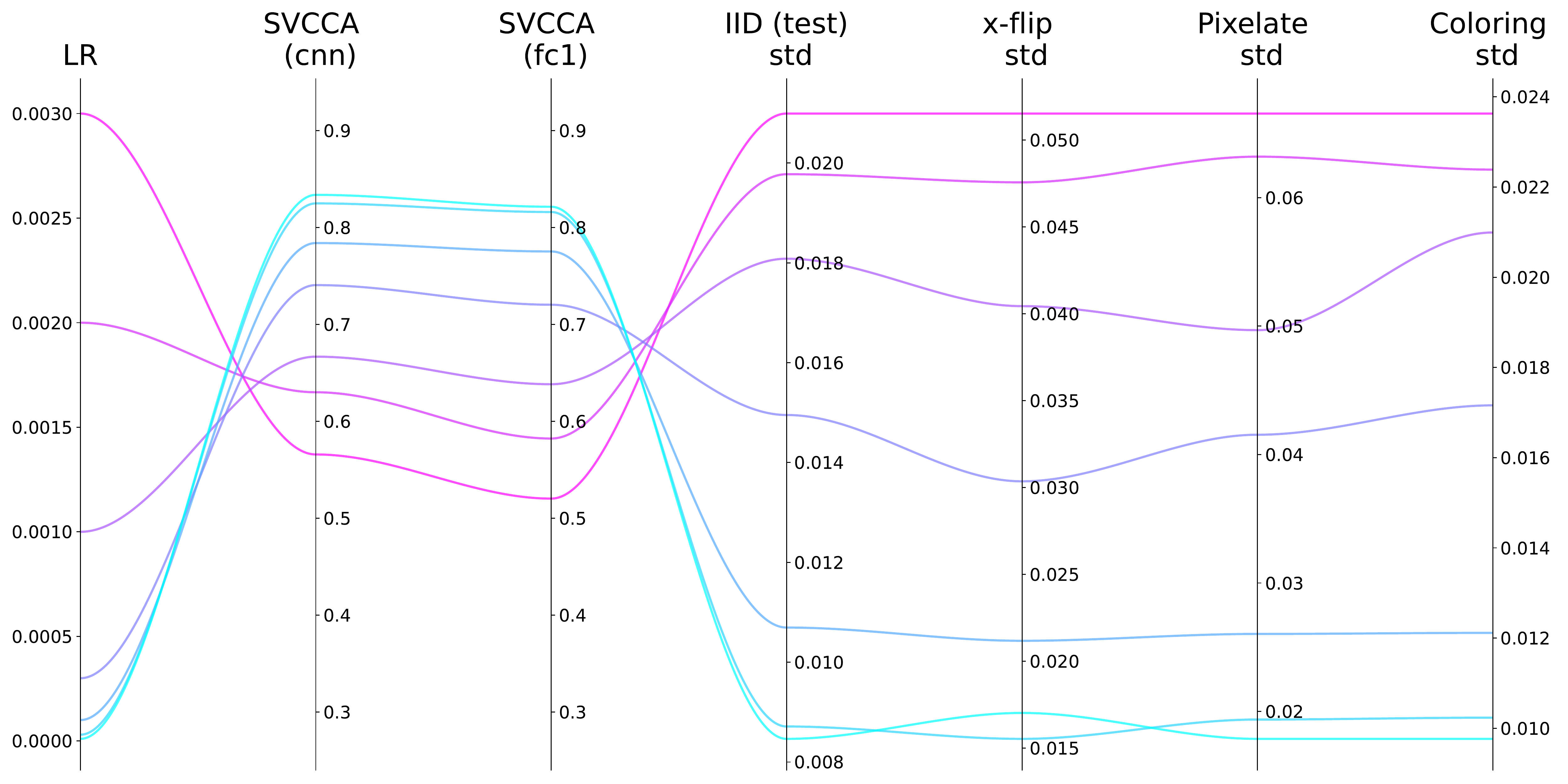}
\includegraphics[width=0.48\textwidth]{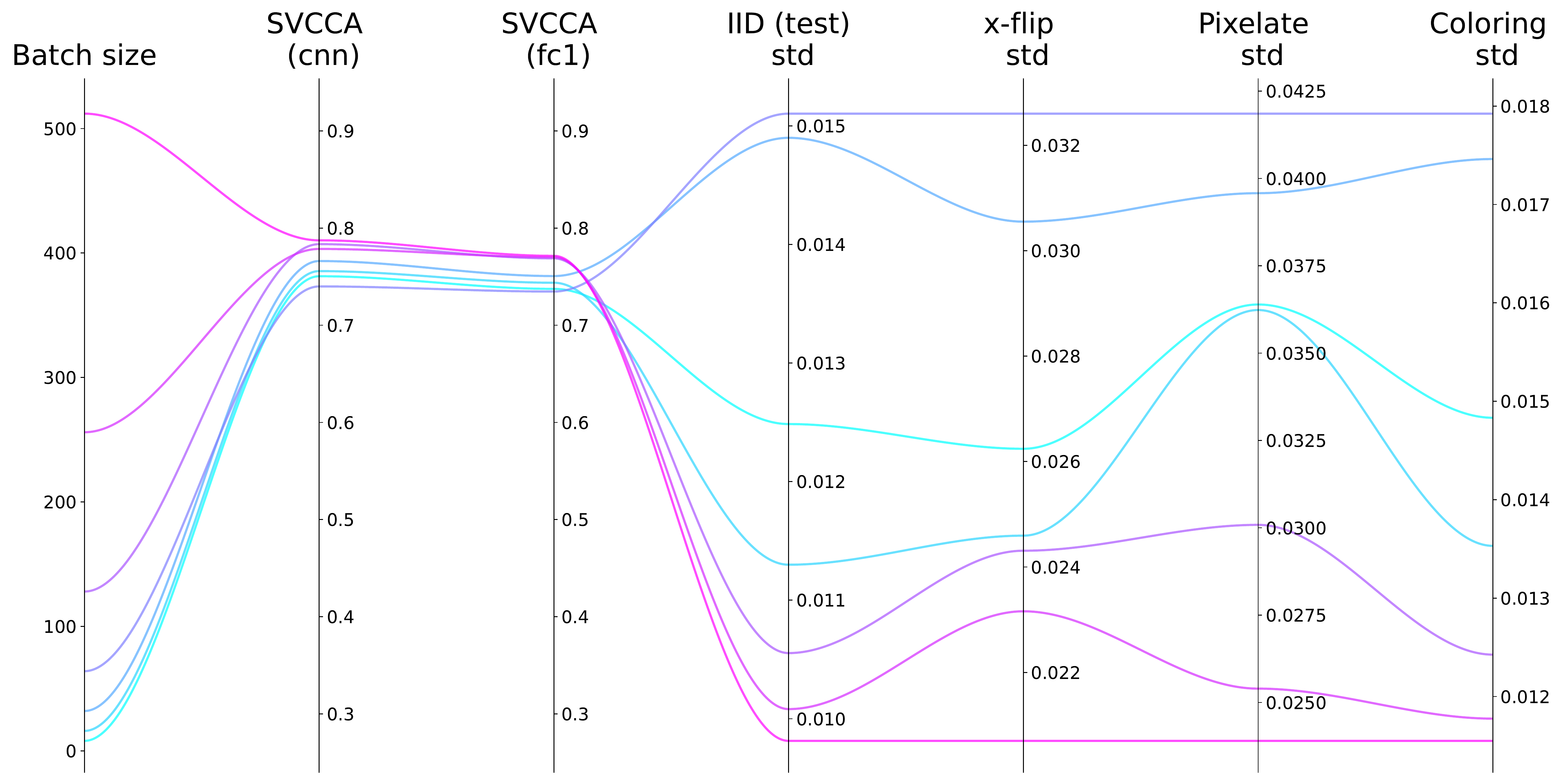}
\begin{tikzpicture}[inner sep=0]
    \node[rotate=90,minimum width=.99\figurewidth,inner sep=1pt,rounded corners=1pt] at (\figurewidth,.55\figurewidth) {\bf \scriptsize CIFAR-9};
\end{tikzpicture}
\includegraphics[width=0.48\textwidth]{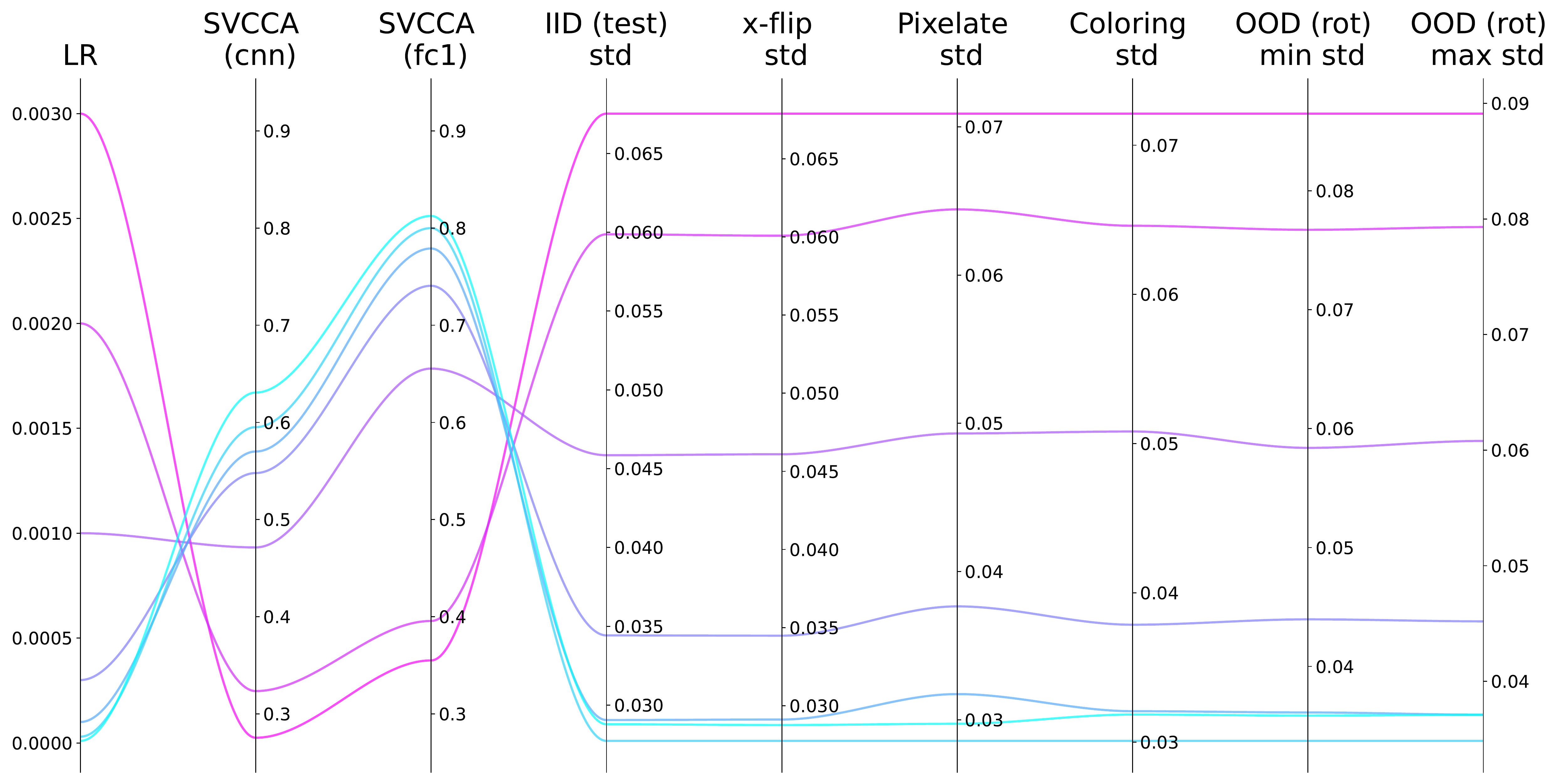}
\includegraphics[width=0.48\textwidth]{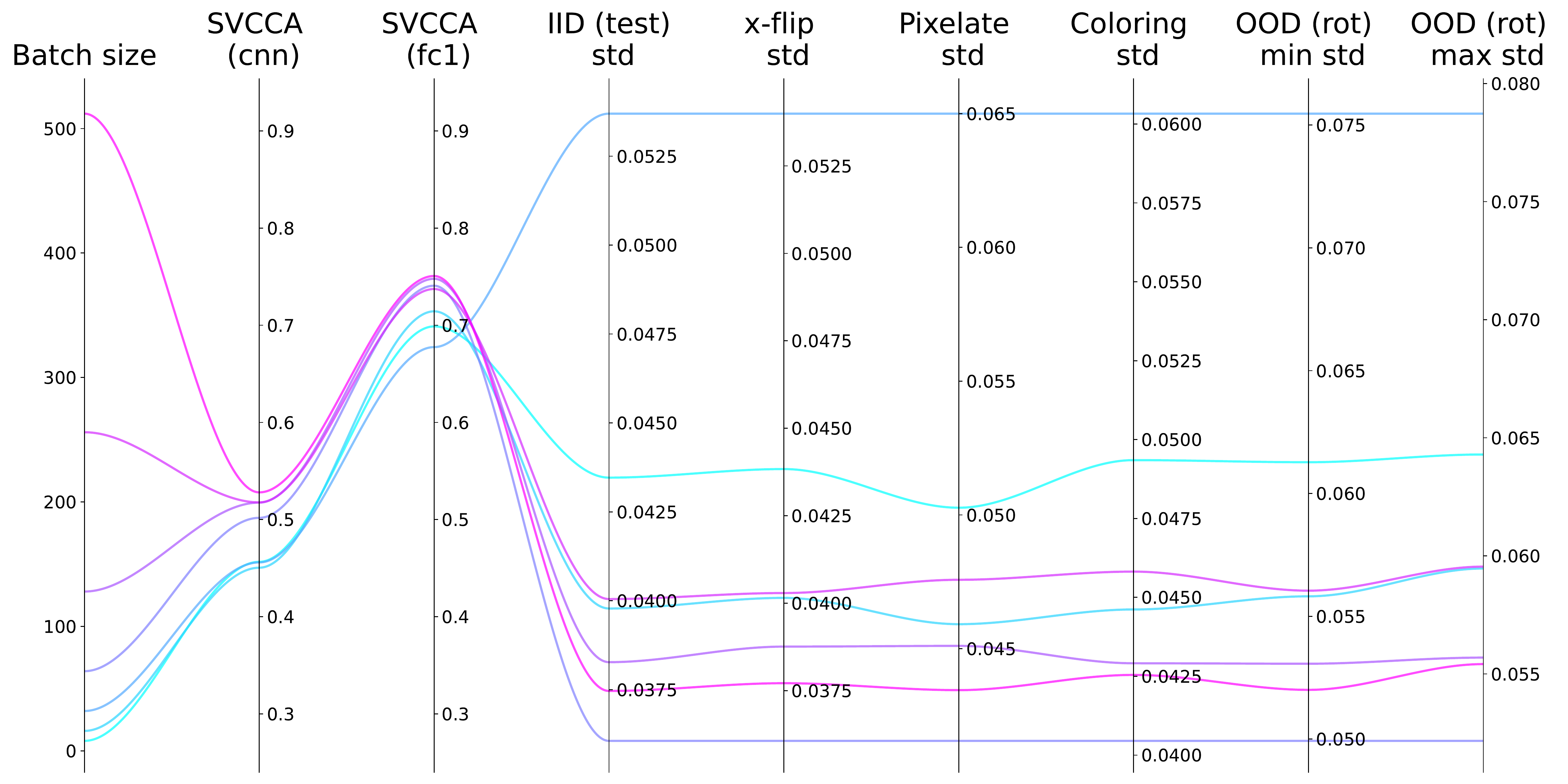}
 \begin{tikzpicture}[inner sep=0]
    \node[rotate=90,minimum width=.99\figurewidth,inner sep=1pt,rounded corners=1pt] at (\figurewidth,.55\figurewidth) {\bf \scriptsize MNIST};
\end{tikzpicture}
\includegraphics[width=0.48\textwidth]{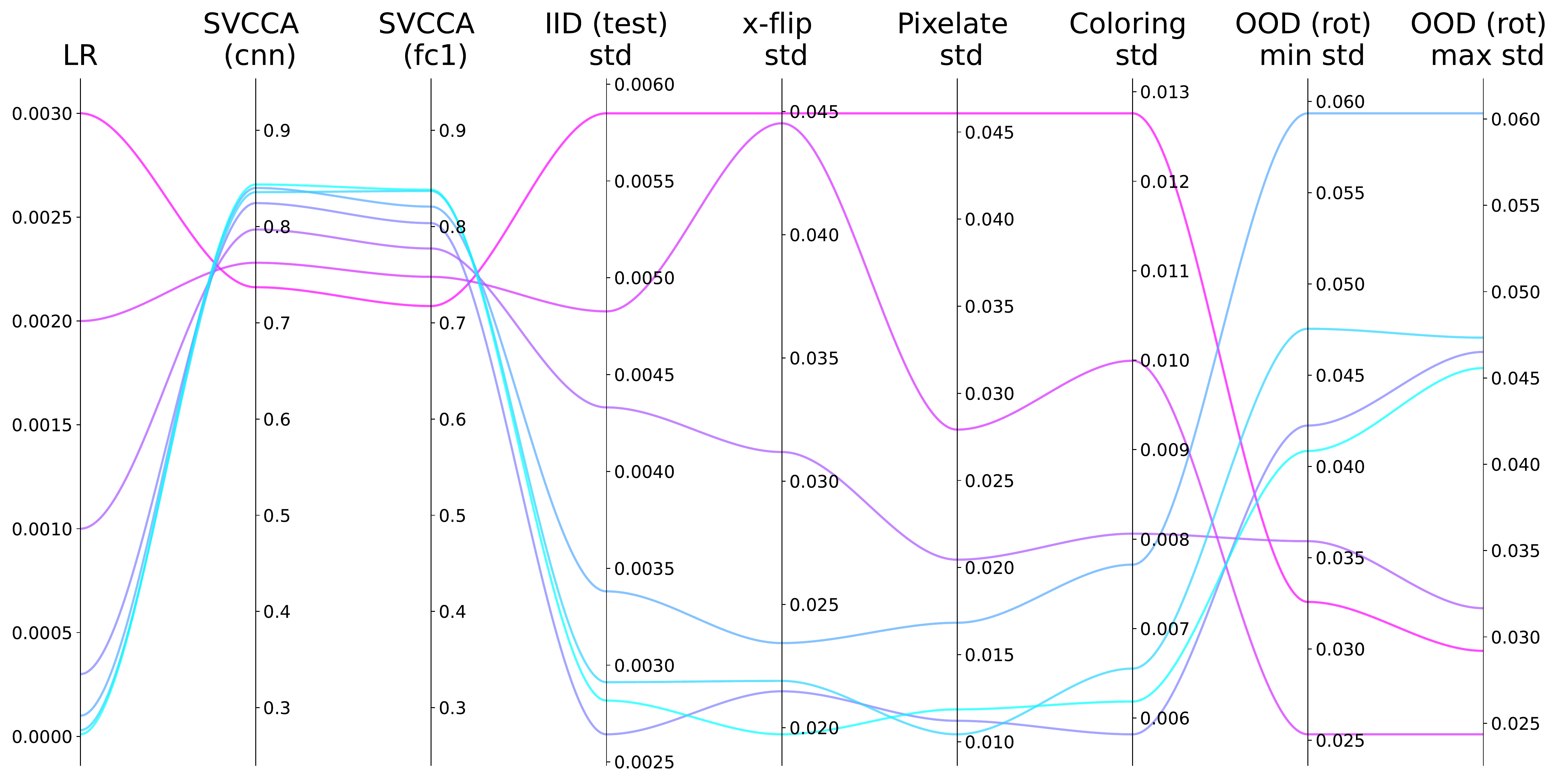}
\includegraphics[width=0.48\textwidth]{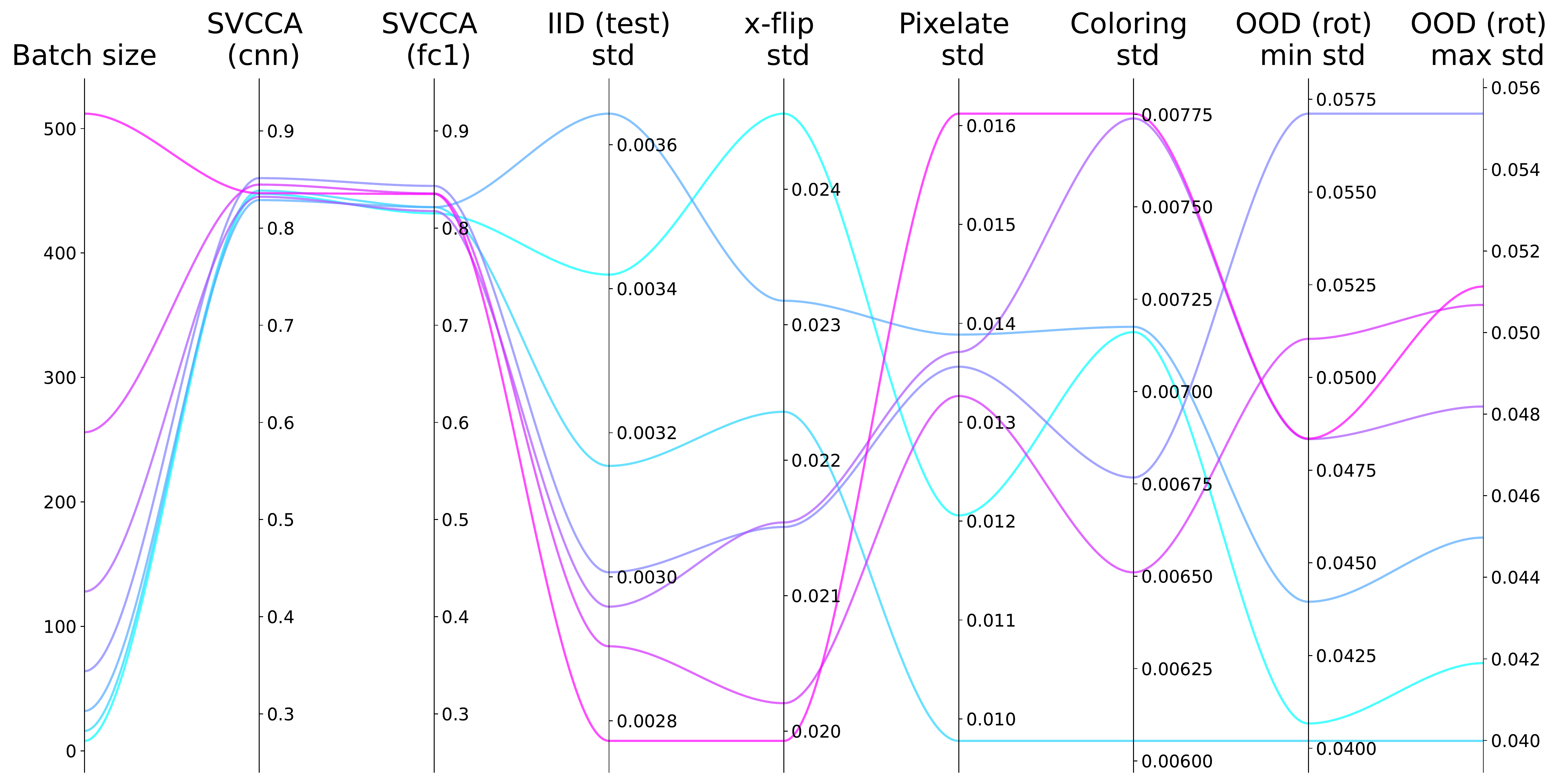}
  \caption{Associating representative (RM) and predictive (PM) multiplicity in two hyper-parameter regimes (\textbf{Left:} learning rate, \textbf{Right:} batch size) for three data sets, with identical model architectures. SVCCA is the measure of inverted RM at two feature layers, while the prediction `std' columns measure PM for the i.i.d. and various OOD distributions. data sets: FashionMNIST (top), CIFAR-9 (middle), and MNIST (bottom). Low learning rate and high batch size correlate with higher SVCCA (lower RM) and smaller variance (lower PM). Thus, across the datasets, RM has a peculiar yet predictable dependence on LR and batch size. \label{fig:hp}}
  \vspace*{-3pt}
\end{figure*}

\section{Methods}
\label{methods}

In order to illustrate representational multiplicity from various perspectives, we now proceed to find a sufficiently precise definition for it and make it measurable. We then evaluate its sensitivity to seemingly irrelevant hyper-parameters of the training strategy under a fixed model. Such hyper-parameters could involve training batch size, learning rate, initialization, \etc. Finally, we show how to disentangle the predictive power of RM and PM.

\paragraph{Problem Setup} We start with the classical setup for a supervised prediction problem, in which one aims to learn a function $h$ that 
maps inputs $\vx \in \X$ to targets $y \in \Y$, \ie, $h \colon \X \to \Y$.
To learn $h$, one minimizes the empirical risk $E(\ell(\vx_i, y_i))$ based on a training set $\MX_{\text{train}} = \{\vx_i\}^N_{i=1}$, $\vy_{\text{train}} = \{y_i\}^N_{i=1}$.
To leverage $h$ as a predictor on a final deployment target system, one usually assumes the test data $\MX_{\text{test}}$ to be drawn from the same distribution and satisfy the common i.i.d. assumption.

\paragraph{Predictive Multiplicity}
Let $H = \{h_k\}^K_{k=1}$ be a set of classifiers such that the difference in empirical risk $|E(h_k) - E(h_j)| < \epsilon$ for all $h_k,h_j \in H$ with $k \neq j$ and a given error tolerance $\epsilon > 0$. 
We define Predictive Multiplicity (PM) over a data set $S$ and $H$ as:
\begin{equation}
\PM(S, H) = \Ex_{\vx \in S} \left[ \sqrt{\Var_{h\in H} \{ h(\vx)\}} \right].
\label{eq:pm}
\end{equation}
In words, we measure the extent to which the risk-equivalent model variants assign conflicting predictions over the data set. Note that while we follow the concept definition of \citet{marx2020predictive}, we diverge from their mathematical definition.

\paragraph{Representational Multiplicity} Representational multiplicity can be defined in terms of three constraints. \textbf{(1)} It measures the variation across the internal representations of risk-equivalent models. This is in contrast to the variation in \textit{predictions}. The representations are expressed by activations in the function space. Unlike evaluation in the weight space, the (function) activation space yields a similarity estimate informed by the available data. Hence, we focus on dissimilarities across the learned functions rather than across their parametrizations. The latter can be diverse even if the underlying functions are not \citep{garipov2018loss}. \textbf{(2)} The multiplicity and risk must be defined in terms of specific data, typically the training or testing data already available in the context. \textbf{(3)} The relevant variation across the representations must be asymmetric in terms of trivial transformations such as affine transformations. This prevents us from, \eg, treating the different order of the same weights in latent space as different neural network representations. Importantly, this excludes the use of straightforward measures such as standard deviations in the activation space. Still, the notion of what constitutes a `trivial' transformation leaves room for interpretation. Here, we only consider the invariance of the output activations to linear transformations. Thus construed, without exhausting every possible trivial transformation, \textit{one can only ever arrive at a lower bound for similarity and upper bound for RM} (also see \citet{ainsworth2022} for possible deeper symmetries).

A suitable approach for this purpose is the singular vector canonical correlation analysis (SVCCA) metric \citep{raghu2017svcca}, a method for assessing activation similarity via combining  Singular Value Decomposition with Canonical Correlation Analysis \citep[CCA,][]{hardoon2004canonical}.

With SVCCA, each neuron $j =1,2,\dots,M$ is represented as a vector of its \textit{output activations} $\vz_j(\vx) = [z_j(\vx_1), \dots, z_j(\vx_N)]^\top$ across all the data points. We then compare a single layer of $M$ neurons between networks 1 and 2 as matrices $\MZ_1 \in \R^{M \times N}$ and $\MZ_2 \in \R^{M \times N}$ composed of the output activation vectors.
This setup satisfies the criteria \textbf{(1)} and \textbf{(2)} mentioned above.
To run the comparison, SVCCA first performs SVD on $\MZ_1$ and $\MZ_2$ and picks the subspaces $\MZ^{'}_1  \subset \MZ_1$ and $\MZ^{'}_2  \subset \MZ_2$ to explain the desired degree (\eg, $99\%$) of the variance in each subspace. Finally, one runs the CCA to find a linear transformation of each subspace
so as to maximize the correlations $\rho_i$ between them. The average of the (maximized) correlations is, then, the SVCCA similarity.

For a single metric for representational multiplicity, we average the largest correlations. It was observed in \citet{raghu2017svcca} that taking the top $25$ vectors (we found $20$ sufficient) accounted for almost all the variance across image data sets, allowing us to discard the rest. Finally, defining multiplicity as the negated expectation over correlations, we arrive at representational multiplicity given as
\begin{equation}
\RM(S,f_1,f_2) = -\frac{1}{T}\sum_{i=1}^T\rho_i(f_1,f_2) ,
\end{equation}
s.t.\ $\rho_1(f_1,f_2) \geq \rho_2(f_1,f_2) \geq \dots$ for a single similarity measure of data set $S$ and (sub-)networks $f_1$ and $f_2$ with SVCCA correlations $\mathbf{\rho}({f_1,f_2})$.
Since CCA is invariant to affine transformations, it satisfies the criteria \textbf{(3)} above, making the method well-suited for RM estimation.
For SVCCA among the comparison set of $K>2$ network variants, we paired every variant $u$ with every other and took the mean of the triangular matrix produced by all pairings:
\begin{equation}
\RM(S, \mathbf{f}) = {K\choose 2}\sum_{u=1}^{K}\sum_{v = 1}^{u} \RM(S,f_u,f_v).
\label{rmeq}
\end{equation}
We considered SVCCA separately for the last CNN layer (`cnn') and the first fully connected one (`fc1').

\paragraph{Assessing Patterns in RM and PM} One can now assess the dependency of RM and PM on hyper-parameters as follows. First, for a given choice of hyper-parameter (`training strategy'), we need to train $K$ variants of our target models, changing nothing but the random seed, yielding variants $h_{k} \in H$, $k=1,2,\dots,K$. We stop the training in each case once we reach the predictive accuracy that we presume can be achieved by all the variants and all the hyper-parameter values. This typically requires extra training runs.
Second, for each variant $h_k$, we need to look at the {\em individual} predictions for each $\vx_i$ in the data set. This yields the variance of predictions across the variants $h_k$ of $H$ and hence PM (\cref{eq:pm}). Third, we do the same for the activations and apply SVCCA to get the RM (\cref{rmeq}).
Finally, we select the hyper-parameter we wish to evaluate (`regime'), and repeat the steps above for $M$ different values of it (`training strategies'). Within each regime, we track whether one strategy yields higher PM or RM than another (\cref{fig:hp}).

\paragraph{Irreducibility of RM}
Computing RM is relatively expensive, while PM can be computed directly from inferred outputs. This raises the question of whether we could just use PM everywhere. But as we have seen earlier, for a finite sample, we should expect that some samples will exhibit RM without changing model outputs and, therefore, without causing any observable PM. In other words, if the subset is not large enough, the PM of the subset does not predict the PM of the full data set. However, since RM is not a function of PM, and RM is expected to be predictive of PM, RM over the subset should predict the PM over the full data set. Essentially, we want to express that for some subsets of data, the RM over the subset is a better predictor of the `real' PM (measured overall data) than PM over the subset. This can be formalized as the following hypothesis.
\pagebreak

{\em Hypothesis 1}. Given data set $S$ of size $N$, consider a subset $s_i \subset S$ with size $N' \ll N$, and $K$ model variants $h_{ij}$ with $j=1,\ldots,K$ for each training strategy $H_i$. Define the estimation errors
$\mathcal{E}_{RM}(s_i, S, H_i) = [c \cdot RM(s_i,H_i) - c' \cdot PM(S, H_i)]$ and
$\mathcal{E}_{PM}(s_i, S, H_i) = [c'' \cdot PM(s_i,H_i) - c' \cdot PM(S, H_i)]$,
with constants $c$, $c'$, and $c''$ scaling each type of a measure to 1 ($c = 1/\max_{i}(RM(s_i,H_i))$, \etc). Now, for each $i$, we can pick $s_i$ in a way that is independent of RM measures, such that the expectation over $L$ training strategies is $\Ex_{i=1,\ldots,L}{\mathcal{E}_{RM}(s_i, S, H_i)} < \Ex_{i=1,\ldots,L}{\mathcal{E}_{PM}(s_i, S, H_i)}$.

\begin{figure*}[!t]
  \begin{subfigure}[t]{.48\textwidth}
    \centering
    \includegraphics[width=\linewidth]{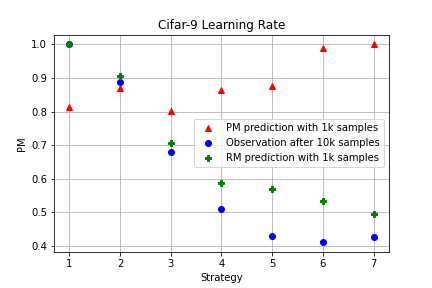}
  \end{subfigure}
  \hfill
  \begin{subfigure}[t]{.48\textwidth}
    \centering
    \includegraphics[width=\linewidth]{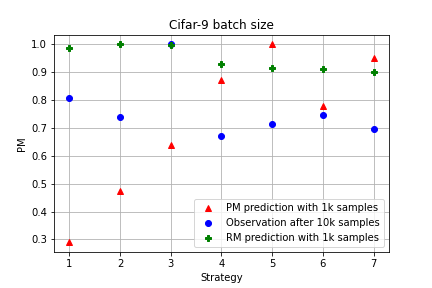}
  \end{subfigure}

  \begin{subfigure}[t]{.48\textwidth}
    \centering
    \includegraphics[width=\linewidth]{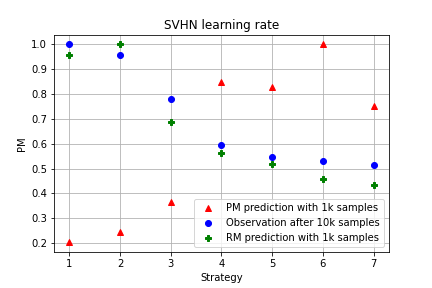}
  \end{subfigure}
  \hfill
  \begin{subfigure}[t]{.48\textwidth}
    \centering
    \includegraphics[width=\linewidth]{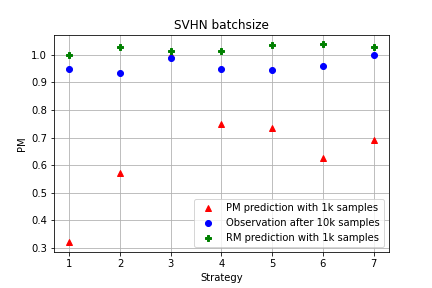}
  \end{subfigure}%
  \caption{Relevance of RM. We compare the predictive power of RM vs \ PM in two data sets under two hyper-parameter regimes, each. In order to show that there \textit{exist} cases where RM cannot be reduced to PM, we first select a subset of 1000 samples for each data set such that those samples have the lowest overall PM. For each strategy (descending learning rates: $0.003, 0.002, 0.001, 0.0003, 0.0001, 0.00003, 0.00001$ or increasing batch sizes: $8, 16, 32, 64, 128, 256, 512$), we then plot the PM of those 1000 samples for each strategy (red, PM-1k), the actual final PM after observing all 10k or 26k samples (blue), and the RM computed from the same subset of 1000 samples (RM-1k). This comparison surfaces the strong correlation between RM-1k and the PM of the full data set, whereas the PM-1k and the PM of the full data set are uncorrelated. This confirms the intuition that, over a small sample with little observed PM, RM can be superior to PM in predicting the final PM that will become observable with a larger sample.\looseness-1}
  \label{fig:pmrm}
  \vspace*{-3pt}
\end{figure*}

If the hypothesis holds, then for RM and PM over a finite sample size, it is possible for the RM to be a better predictor of the full set PM. Hence, RM is not reducible to PM. In order to examine the hypothesis, we can pick $s_i$ in the most obvious way, by taking the samples that minimize PM over $H_i$. %
As this sample-picking method is uninformed by RM, it serves as proof by example that a finite sample {\em can} allow such situations to occur.
We show that in all data sets, the hypothesis clearly holds (\cref{fig:pmrm}). %

\paragraph{Arbitrary Predictions}
In the traditional best-model based analysis of classification models, it is common to review the samples most prone to classification errors \textit{for that model}, often presented as a confusion matrix. Now, admitting the possibility of RM, one can similarly ask which samples are most prone to being {\em differently classified} \textit{across various equally accurate variants of the model}. This yields a sort-of `second-order' confusion analysis that we call {\em confabulation analysis}, visualized in terms of the inputs that yield the highest multiplicity values (\cref{fig:confab}).
The terminology is motivated by the fact that the choices the model makes for those samples appear arbitrary, or {\em confabulated}, irrespective of how confident a specific model instance is about them. %

There is a major conceptual difference between confusion and confabulation. Confusion informs us which kind of samples our model is \textit{currently wrong about}, suggesting we should do something about it. Confabulation informs us which samples our models might be \textit{fundamentally unable to classify correctly}, so ambivalent or devoid of relevant information that they should perhaps be discarded all together.
In a disturbing contrast, the common approach to force one's model to classify even the worst of these samples `correctly' may be akin to learning to classify noise.

\section{Experiments}
\label{results}

We conduct three types of experiments to support our case. First, we relate RM, PM, and OOD generalization in various data sets, under various training scenarios. Second, we show hypothesis 1 to hold on all the examined data sets, showing that one cannot simply reduce RM to PM. Third, we visualize the concrete manifestations of high RM.

\paragraph{Data Sets and Architectures}
We examined RM on classifiers trained on four typical small vision data sets: MNIST, FashionMNIST \citep{xiao2017fashionmnist}, CIFAR-9 (CIFAR-10 \citep{cifar10} with one dropped class), and SVHN \citep{netzer11}. We used SVHN as an OOD data set for MNIST and STL9 (STL10 \citep{stl10} with one dropped class to match CIFAR-9) as an OOD data set for CIFAR-9. Our focus was on typical scenarios, therefore, we picked a simple and identical CNN architecture for each data set and controlled for the same acceptable, but not state-of-the-art, accuracy within the data set (see \cref{app:B} for training and architecture details).

\paragraph{Peculiar Regularities in Training Strategies}
We focused on relevant measurable differences in predictions and representations of risk-equivalent (\ie, equally accurate) models trained with the same data, model structure, and hyper-parameters. The baseline assumption was that such models should behave equivalently, despite hyper-parameter differences during training. %
To measure the empirical divergence from this assumption, for each strategy, we first fix all hyper-parameters and train 10 variants while only changing the random seed. We measure the SVCCA similarities and output variance, separately for manual augmentations that transform the data set to OOD and for the matching separate OOD data sets. In \cref{fig:hp}, a single continuous line represents a single strategy. We then repeat the measurements for 7 values of each hyper-parameter, separately for batch size and learning rate. This yields $2 {\times} 7 {\times} 10$ training runs per data set (the 7 continuous lines).

To ensure that the results within each data set are risk-equivalent, we first had to determine the maximum (test set) accuracy that can be reached using any of the hyper-parameter values in the tested range. For the actual training sessions, we then stopped the training in each case at this value. Following the training to the highest achievable accuracy would not have made a difference to the conclusions (see \cref{app:C}).

For each line, one should look for any suspiciously small SVCCA (high RM) and large i.i.d. test set prediction variance (high PM) values, whether the two are correlated, and how they relate to the OOD prediction variances. %
Across the lines, the first interesting question is how widely dispersed they are. Any changes in their SVCCA values reflect the effect of the hyper-parameter on the multiplicity of potential solutions. Finally, the correlation between the varied hyper-parameter and the magnitude of the ensuing PM or RM is examined.
Across the data sets, the altered hyper-parameters correlate with RM. The larger batch size and the smaller learning rate correlate strongly with higher SVCCA in both examined layers (\cref{fig:hp}) and with smaller PM both in the i.i.d.\ and OOD test sets. (See also \cref{fig:parallel2} in App.)

Of the four data sets included, the only distinctly different behaviour was observed for MNIST under the {\em batch size strategy}. This is likely due to the small range in SVCCA values, especially since MNIST under the {\em LR strategy}, with a wide SVCCA range of $[0.737, 0.849]$, shows the same regularities as observed with other data sets (\cref{fig:hp}). As an alternative to SVCCA, we also attempted the CKA metric \citep{hinton2018} for the same purpose, but failed to observe any correlation between CKA and PM values, hence we chose not to use it for the analysis.
Measurements of Pearson correlation coefficient (\cref{sample-table}) across data sets support the same conclusions.

\begin{figure*}[!t]
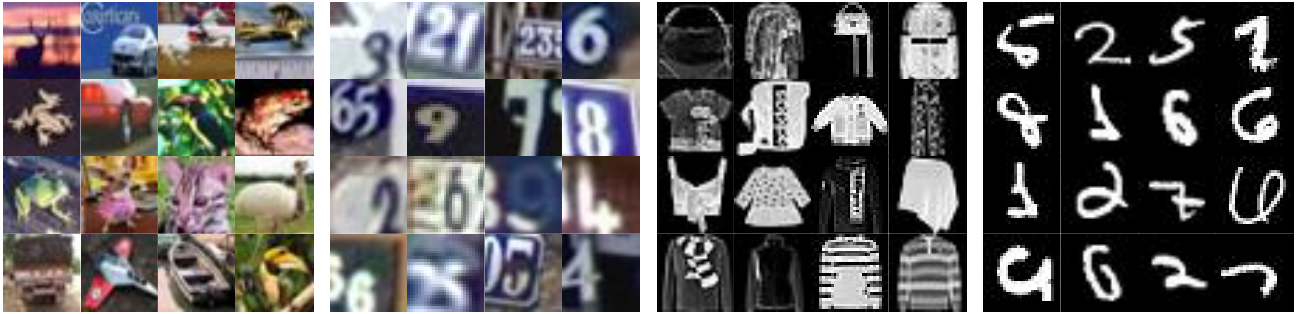

    \centering\scriptsize
    \setlength{\figurewidth}{.060\textwidth}
    \setlength{\subfigurewidth}{.24\textwidth}
    \setlength{\figureheight}{\figurewidth}
    \begin{subfigure}{\subfigurewidth}
    \begin{tikzpicture}[inner sep=0]
      \foreach \x [count=\i] in {0,...,15}
        \node[draw=white,fill=black!20,minimum size=\figurewidth,inner sep=0pt]
          (\i) at ({\figurewidth*mod(\i-1,4)},{-\figureheight*int((\i-1)/4)})
        {\includegraphics[width=\figurewidth]{./fig/confab_cifar9_arch4_batchsize_\x.jpg}};
    \end{tikzpicture}
    \end{subfigure}
    \hfill
    \begin{subfigure}{\subfigurewidth}
    \centering\scriptsize
    \begin{tikzpicture}[inner sep=0]
      \foreach \x [count=\i] in {0,...,15} %
        \node[draw=white,fill=black!20,minimum size=\figurewidth,inner sep=0pt]
          (\i) at ({\figurewidth*mod(\i-1,4)},{-\figureheight*int((\i-1)/4)})
        {\includegraphics[width=\figurewidth]{./fig/confab_SVHN_arch2_batchsize_\x.jpg}};
    \end{tikzpicture}
    \end{subfigure}
    \hfill
    \begin{subfigure}{\subfigurewidth}
    \centering\scriptsize
    \begin{tikzpicture}[inner sep=0]
      \foreach \x [count=\i] in {0,...,15} %
        \node[draw=white,fill=black!20,minimum size=\figurewidth,inner sep=0pt]
          (\i) at ({\figurewidth*mod(\i-1,4)},{-\figureheight*int((\i-1)/4)})
        {\includegraphics[width=\figurewidth]{./fig/confab_fashion_arch1_batchsize_\x.jpg}};
    \end{tikzpicture}
    \end{subfigure}
    \hfill
    \begin{subfigure}{\subfigurewidth}
    \begin{tikzpicture}[inner sep=0]
      \foreach \x [count=\i] in {0,...,15} %
        \node[draw=white,fill=black!20,minimum size=\figurewidth,inner sep=0pt]
          (\i) at ({\figurewidth*mod(\i-1,4)},{-\figureheight*int((\i-1)/4)})
        {\includegraphics[width=\figurewidth]{./fig/confab_MNIST_arch1_batchsize_\x.jpg}};
    \end{tikzpicture}
    \end{subfigure} 
    \caption{Top-16 confabulation inputs for CIFAR-9, SVHN, FashionMNIST, and MNIST. Confabulation measures the diversity of predictions assigned to each sample, across different training runs of the same model. In other words, the highly confabulated items are those that will most likely be classified differently under different runs. This measure is more informative than the aggregate classification error. Considering some of these results, it seems that forcing the model to learn their `correct' class amounts to learning noise.}
  \label{fig:confab}  
  \vspace*{-3pt}
\end{figure*}

\paragraph{Relevance of RM as Predictor} Given the correlations between RM and PM, one might well ask, do we really need RM in the first place, when PM often appears to be more easily accessible? Conceptually, RM must necessarily precede PM, but at the limit of infinite test data, all the interesting variation in RM must necessarily be perceived as PM. On the other hand, under limited data, some of the internal variation (RM) will still be collapsed at the final layers, at least under operations such as softmax activation. Hence, following Hypothesis 1 in \cref{methods}, we demonstrate the difference empirically, by constructing the \textit{existence} proof for a case where the two are not equivalent.

To this end, we first measured PM for the complete training data sets of Cifar-9 (10,000 samples) and SVHN (26,000 samples). Then, we picked a subset of 1000 samples with the lowest PM, as these items are likely to yield the least amount of information about RM. Using the same samples, we measured RM. Finally, we asked, is the PM or RM in the 1k sample predictive of the PM for the \textit{whole} data set, when measured for different training strategies? As seen in \cref{fig:pmrm}, after scaling all values to $[0,1]$ range, the RM values for the 1k subset are strongly correlated with the PM values of the whole data set, whereas the PM of the 1k test set is uncorrelated. Note that the 1k sample was in no way selected to favour the utility of RM values as such. Hence, in each case, we have shown that RM captures information that is not reducible to the information captured by PM.

\paragraph{Confabulator Inputs}
We visualize the confabulation matrices in \cref{fig:confab} for all data sets. The confabulators appear similar to what one might expect in confusion matrices, but here, they reflect not just a single training run, but a range of independent sessions. (Note that the confabulations are computed across all training strategies, and could further be examined for each recipe separately.)

\begin{table}[t!]
  \caption{The Pearson correlation coefficient (PCC) between negative representational multiplicity (RM) measured using SVCCA metric, and predictive multiplicity (PM) measured using standard deviation between OOD predictions across the data sets (LR = learning rate). The columns `x-flip' (horizontal flip), `Pixelate' and `Coloring' represent the OOD predictions, followed by an exposition of the range of RM values for the data set and the strategy ($\Delta$~SVCCA). Strong negative correlation values indicate that PM and RM align.} %
  \label{sample-table}
  \centering\scriptsize
  \setlength{\tabcolsep}{1.75pt}
  \begin{tabular}{lcccccccc}
    \toprule
    & \multicolumn{2}{c}{x-flip} & \multicolumn{2}{c}{Pixelate} & \multicolumn{2}{c}{Coloring} & \multicolumn{2}{c}{$\Delta$~SVCCA (fc1)}\\
    \cmidrule(lr){2-3} \cmidrule(lr){4-5} \cmidrule(lr){6-7} \cmidrule(lr){8-9}
    Data set & batch size & LR & b.\ size & LR & b.\ size & LR & b.\ size & LR\\
    \midrule
    F-MNIST & $-.750$ & $-.991$ & $-.872$ & $-.985$ & $-.767$ & $-.978$ & $.037$ & $ .301 $\\
    CIFAR-9  & $-.623$ & $-.990$ & $-.571$ & $-.980$ & $-.609$ & $-.985$ & $ .073$ & $ .445 $\\
    SVHN  & $-.422$ & $-.986$ & $.191$ & $-.985$ & $.089$ & $-.985$ & $.033$ & $.234$\\
    MNIST & $-.482$ & $-.977$ & $.221$ & $-.933$ & $.063$ & $-.932$ & $.041$ & $.167 $\\
    \bottomrule
  \end{tabular}
\end{table}

\section{Conclusion and Limitations}
\label{conclusion}

We propose that practitioners consider explicitly addressing RM-PM distinction as a routine aspect of any ML problem setup. As a preliminary example, considerable representational and predictive multiplicity appeared in all four data sets studied. Within the network architectures considered, certain choices of basic training hyper-parameters appear to correlate with higher occurrence of multiplicity. Given the apparent ubiquity of these phenomena, we suggest that, in general, model evaluation should either include such measures or explicitly address their absence, so as to not overstate the power of the resulting models.

The prevalence of RM raises questions about the ML practises and discourse based on the one-dimensional notion of `the best' model for the data, and its equally imaginary counterpart `unbiased model'. Thus construed, `the best' appears as if transcending the model--data gap altogether and, in some unconditional sense, becoming one with the objective patterns in the data, even in cases of large models and data sets. %
The inevitable failure to really close the gap then leaves room for \emph{ethical} questions in the form of \emph{biases} that cannot be resolved, as if RM had resulted from procedural malpractice.

The experiments rely on the correlations across the top 20 SVCCA vectors as a proxy for RM. While the observed regularities lend credence to this decision, one could search for better proxies for RM. CCA-based methods are limited by their reliance on linear transformations, and do not support comparing different topologies or initialization conditions. For the latter, methods such as centred kernel alignment (CKA) might be a better fit \citep{hinton2018}.

In terms of computational demands, the training time of models dominates. SVD computation in SVCCA, for an $m \times n$ matrix, does have runtime complexity of $\mathcal{O}(m~n~ \min(m,n))$, but remains feasible as the analysis is done on a per-layer basis. To observe trends across several data sets and still ensure reproducibility with the over 500 training runs, we chose to use only low-resolution image data sets. The experiments in classification models could also be repeated for, \eg, generative models. While large image data sets and architectures were not examined, the same principles apply to any parameterized model. Only two training strategies were compared, and we did not study the effects of normalization or different optimizers \citep[ADAM by][was used]{adam}. While these limitations indicate that a larger research effort is called for, our key results are sufficiently consistent across four separate data sets so as to serve as a starting point for follow-up studies.

\clearpage

\bibliographystyle{icml2023}

\clearpage
\appendix
\onecolumn

\section{Extended Results}

\iffalse
\begin{figure}
    \centering
    \input{boxplot}
    \caption{Caption}
    \label{fig:my_label}
\end{figure}
\fi

\begin{figure*}[h!]
\centering
\setlength{\figurewidth}{.23\textwidth}
\begin{tikzpicture}[inner sep=0]
    \node[rotate=90,minimum width=.99\figurewidth,inner sep=1pt,rounded corners=1pt] at (\figurewidth,.55\figurewidth) {\bf \scriptsize FashionMNIST};
\end{tikzpicture}
\includegraphics[width=0.48\textwidth]{parallel_coordinates_fashion_LRconst.pdf}
\includegraphics[width=0.48\textwidth]{parallel_coordinates_fashion_batchsize.pdf}\\
\begin{tikzpicture}[inner sep=0]
    \node[rotate=90,minimum width=.99\figurewidth,inner sep=1pt,rounded corners=1pt] at (\figurewidth,.55\figurewidth) {\bf \scriptsize CIFAR-9};
\end{tikzpicture}
\includegraphics[width=0.48\textwidth]{parallel_coordinates_cifar9_LRconst.pdf}
\includegraphics[width=0.48\textwidth]{parallel_coordinates_cifar9_batchsize.pdf}\\
\begin{tikzpicture}[inner sep=0]
    \node[rotate=90,minimum width=.99\figurewidth,inner sep=1pt,rounded corners=1pt] at (\figurewidth,.55\figurewidth) {\bf \scriptsize SVHN};
\end{tikzpicture}
  \includegraphics[width=0.48\textwidth]{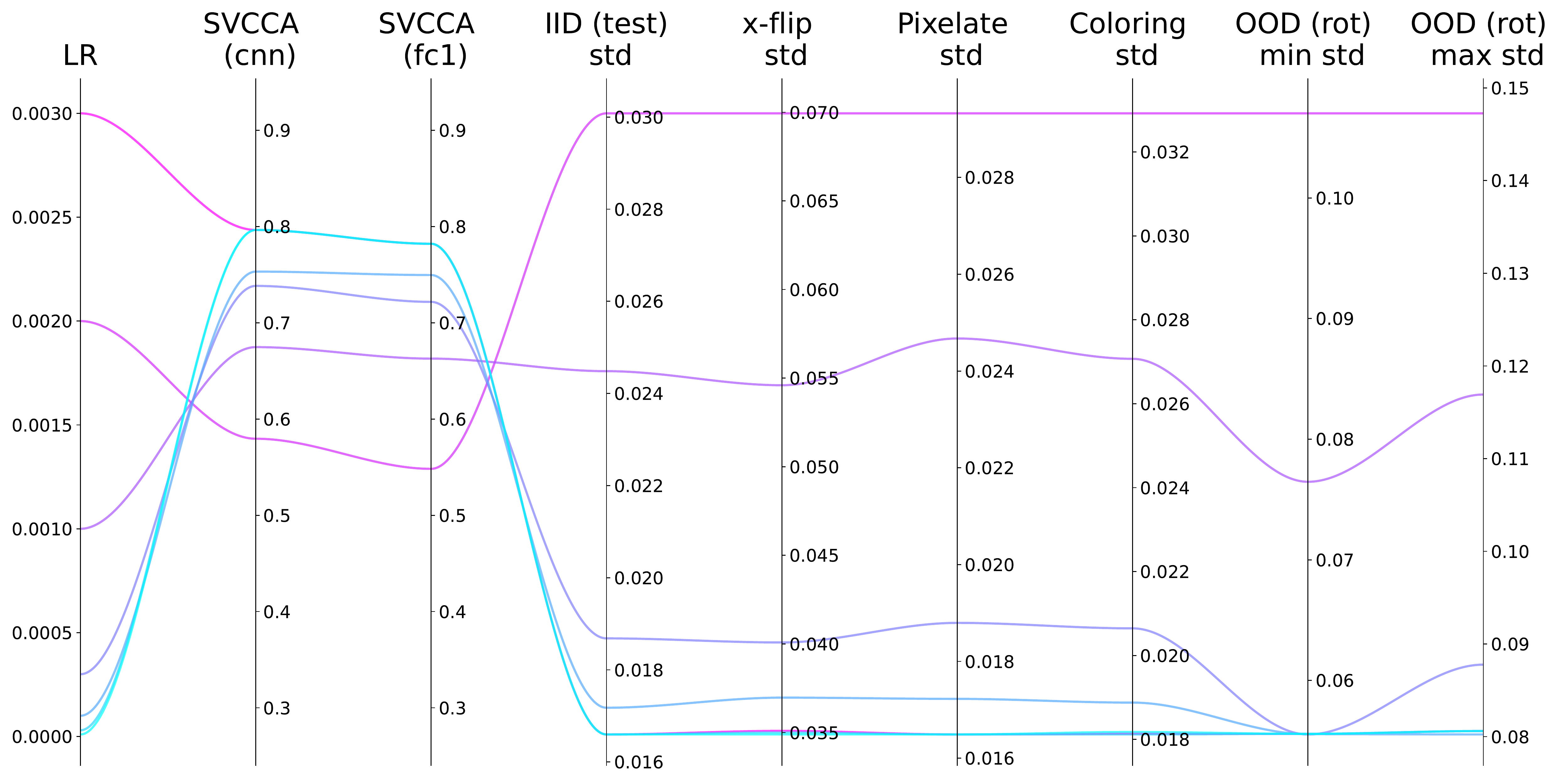}
  \includegraphics[width=0.48\textwidth]{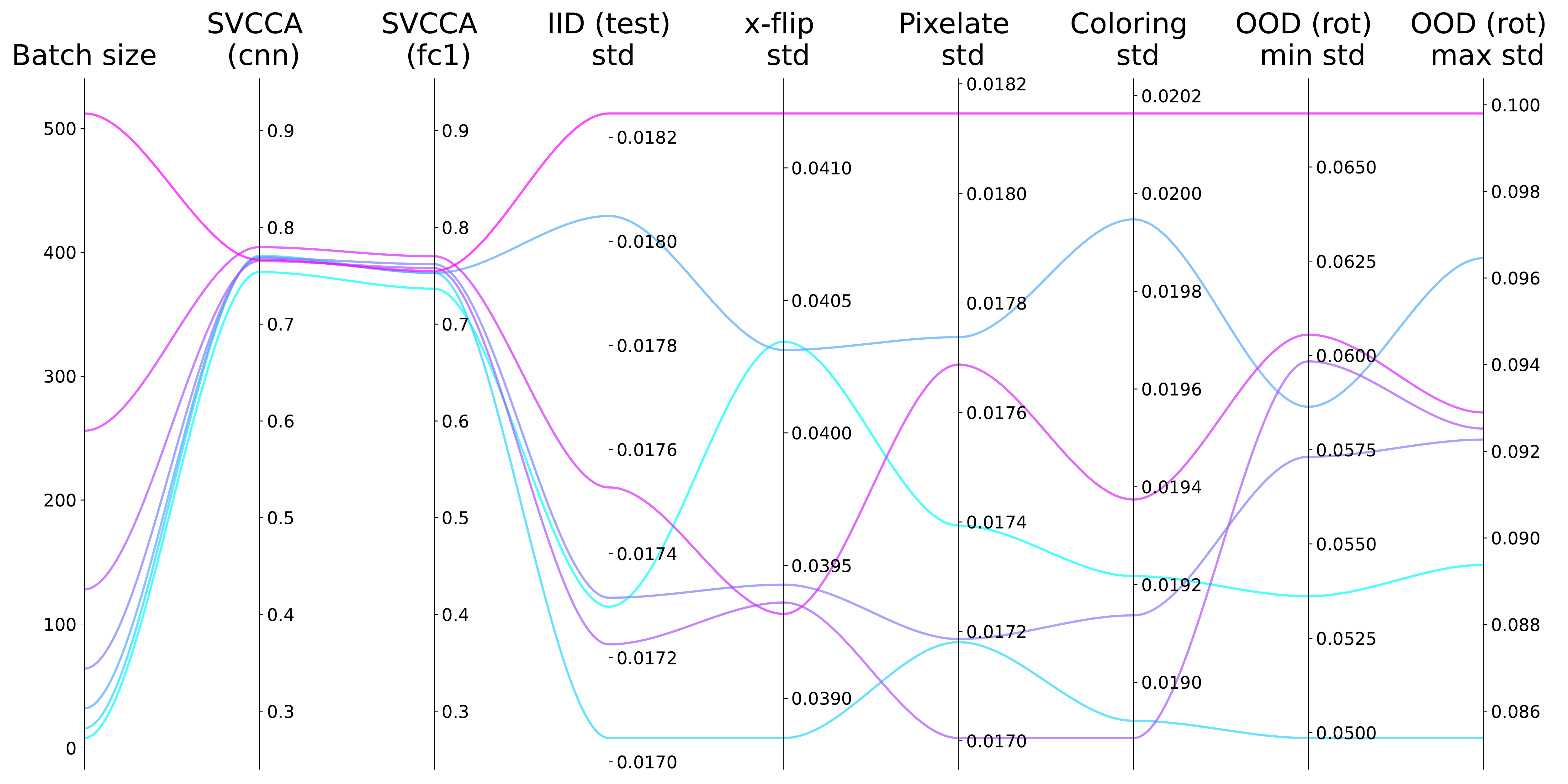}
 \begin{tikzpicture}[inner sep=0]
    \node[rotate=90,minimum width=.99\figurewidth,inner sep=1pt,rounded corners=1pt] at (\figurewidth,.55\figurewidth) {\bf \scriptsize MNIST};
\end{tikzpicture}
\includegraphics[width=0.48\textwidth]{parallel_coordinates_MNIST_LRconst.pdf}
\includegraphics[width=0.48\textwidth]{parallel_coordinates_MNIST_batchsize.pdf}\\
  \caption{
  More comparisons of representative (RM) and predictive (PM) multiplicity, with SVHN data set included to showcase that SVHN in the learning rate regime breaks some of the trends observed with other data sets. Two hyper-parameter regimes are shown (\textbf{Left:} learning rate, \textbf{Right:} batch size) for four data sets. Identical model architectures were used for every hyper-parameter of a data set. SVCCA is the measure of inverted RM at two feature layers, while the prediction `std' columns measure PM for the i.i.d. and various OOD distributions. data sets: FashionMNIST (top), CIFAR-9 (middle), SVHN, and MNIST (bottom). Low learning rate and high batch size strongly correlate with higher SVCCA (lower RM) and smaller variance (lower PM).
  \label{fig:hpall}}
\label{fig:parallel2}
\end{figure*}

\section{Training Details}
\label{app:B}

\paragraph{Architectures } We trained MNIST, SVHN and FashionMNIST with the typical convolutional architecture $conv((1,48,3)-ReLU-MaxPool(2, 2) - conv(48,96,3)-ReLU-MaxPool(2, 2) - conv(96,80,3)-ReLU - conv(80,96,3)-ReLU - FC(96,512)-FC(512,10)$ and Cifar9 with $conv((3,48,3)-ReLU-MaxPool(2, 2) - conv(48,96,3)-ReLU-MaxPool(2, 2) - conv(96,80,3)-ReLU - conv(80,96,3)-ReLU - FC(384,512)-FC(512,9)$.

\paragraph{Training} Apart from the architecture, the same training methods and hyper-parameters were used for each data set and run, with the exception of the batch size and learning rate. In learning rate regime, we used batch size $64$ while using the learning rates $[0.003, 0.002, 0.001, 0.0003, 0.0001, 0.00003, 0.00001]$. In the batch size regime, we used learning rate $0.0001$ and batch sizes $[8,16,32,64,128,256,512]$. Cross entropy loss was used, with ADAM optimizer ($\beta_1=0.9, \beta_2=0.999, \epsilon=10^{-8}$).

For each data set, We first found the maximum test set accuracy comfortably achieved by every strategy for each hyper-parameter regime of that data set. We then trained 10 variants for each of them, varying only the random seed between the runs. We then used the checkpoints at the target levels for our evaluations. The accuracies were reached as in \cref{tbl:acc}. Note that it was not our goal to achieve state-of-the-art accuracy for the experiments of this paper.

\begin{table}
\centering
\caption{Test set accuracy statistics for the data sets used in this work.}
\label{tbl:acc}
\begin{tabular}{l|c}
    \toprule
    Data set (HP regime) & Mean $\pm$ Std \\\midrule
    CIFAR-9 (batch size) & $0.709 \pm 0.001$ \\
    CIFAR-9 (learning rate) & $0.643 \pm 0.005$\\
    FashionMNIST (batch size) & $0.940 \pm 0.007$ \\
    FashionMIST (learning rate) & $0.930 \pm 0.015$ \\
    MNIST (batch size) & $0.979 \pm 0.001$ \\
    MNIST (learning rate) & $0.981 \pm 0.002$ \\
    SVHN (batch size) & $0.905 \pm 0.001$\\
    SVHN (learning rate) & $0.894 \pm 0.002$\\
    \bottomrule
\end{tabular}
\end{table}

\paragraph{data sets} MNIST, FashionMNIST, SVHN, and CIFAR-9 were used for training. STL-9 was additionally used as an extra OOD test data set. CIFAR-9 was made by dropping the class `frog' and STL-9 by dropping the class `monkey', after which the class order of the two sets was made to match so that STL-9 served as the OOD data set for CIFAR-9 trained models. The standard training/validation split was used for each data set. Note that since we explicitly focused on our specific hyper-parameter strategies, there was no need for a separate validation and test data sets, hence, for each data set, the held-out data was treated as the i.i.d.\ `test' data.

For out of distribution (OOD) data sets, we cross-matched data sets as SVHN against MNIST, and CIFAR-9 against STL-9. In addition, other OOD data sets (see \cref{fig:hp}) were created from the original data sets in the following manner. The `x-flip' was created by random horizontal flip with .9 probability, `color jitter' by randomly changing brightness uniformly with a factor $0.0$ to $0.3$ and hue by a factor $-0.1$ to $+0.1$. Pixelation was done by downscaling and upsampling by a factor of 2. 'OOD rot' refers to rotations done to the original data set by uniformly random increments of $0\ldots20$ degrees, $20\ldots30$ degrees, \etc, up to $90\ldots110$ degrees, to yield 10 OOD test cases. The `min std' and `max std' in the figure refer to the smallest and the largest rotation range, respectively.

The data sets contained no person identifiable data. All the data sets allow non-commercial usage for research purposes.

\section{Training to Maximum Accuracy}
\label{app:C}

We take as a given that there are less model variants that achieve the top performance than there are variants that achieve sub-par performance (for empirical comparisons, see Fort 2020). In other words, the increase in validation accuracy should correlate with decreasing diversity of possible solutions.

One should then ask, first, is it possible that the decreased diversity of solutions reached by some hyper-parameter strategies can be explained by the greater achieved accuracy? By controlling for the test accuracy across compared variants (\cref{fig:hp}), we preclude this explanation.

Second, one can further ask, whether the differences could still be explained by the greater \textit{achievable} accuracy, that is, perhaps there is less diversity with the solutions that are on the right path to achieve maximum accuracy? Though the hypothesis appears contrived, we addressed it by checkpointing \textit{but not stopping} the main experiments at the equivalent-risk level. Instead, we also followed each experiment until convergence, and identified the pseudo-maximum accuracy for each training strategy, so that the maximum accuracy was achieved in at least 5 of the runs. (The absolute maximum accuracy would of course, strictly speaking, be only achieved by a single variant, preventing any dispersion analysis.)

In Fig.~5, we show the same results as in Fig.~4, but measured for the variants that achieve the pseudo-maximum accuracy for that strategy. Although one can see some repeating patterns, the decreased diversity of solutions certainly does not correlate with greater achieved accuracy. Hence, it is clear that the variation across the experiments is not explained by differences in either the achieved or potentially achievable accuracy.

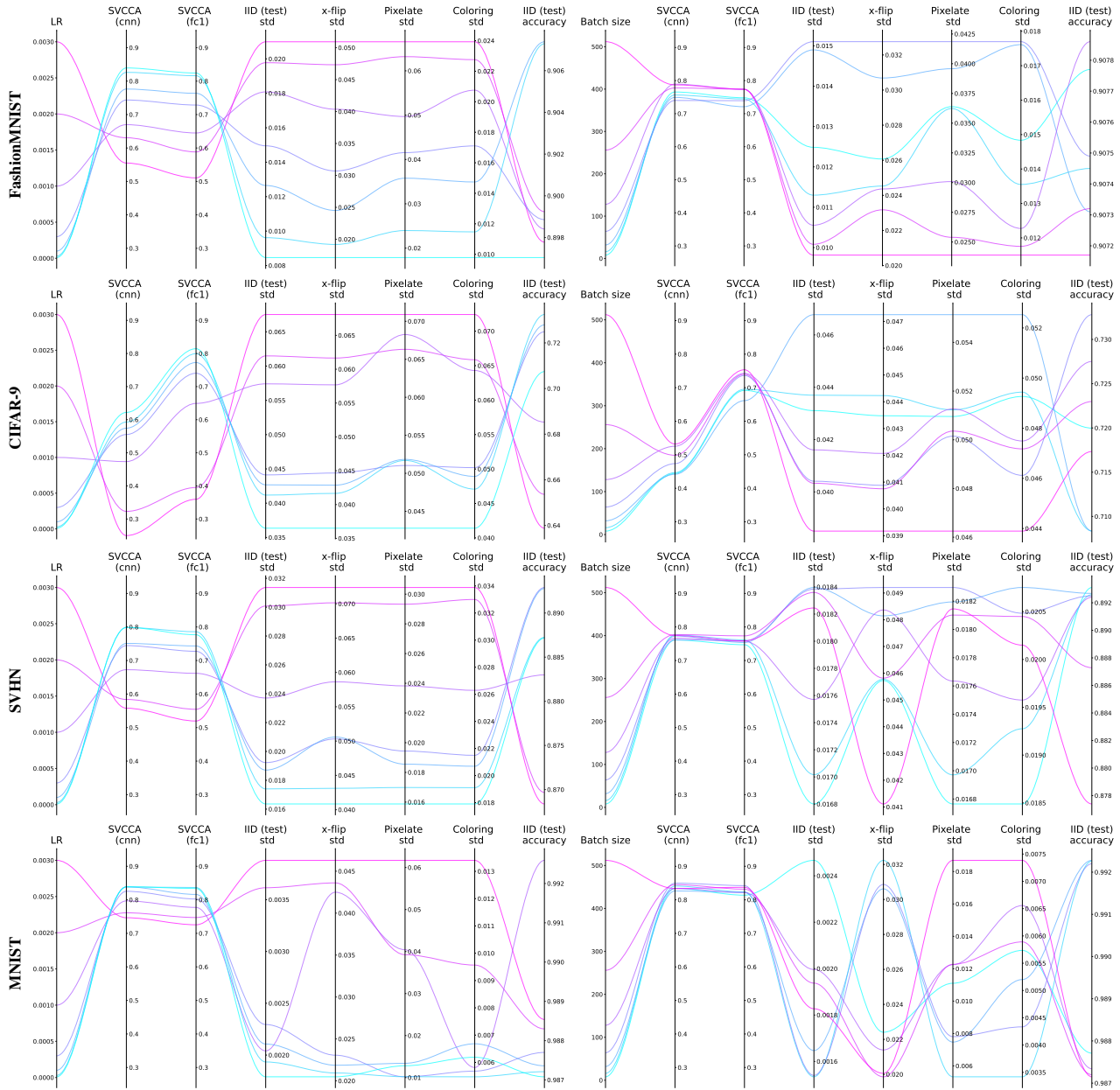
\begin{figure}[h]
\centering
\setlength{\figurewidth}{.23\textwidth}
\begin{tikzpicture}[inner sep=0]
    \node[rotate=90,minimum width=.99\figurewidth,inner sep=1pt,rounded corners=1pt] at (\figurewidth,.55\figurewidth) {\bf \scriptsize FashionMNIST};
\end{tikzpicture}
\includegraphics[width=0.48\textwidth]{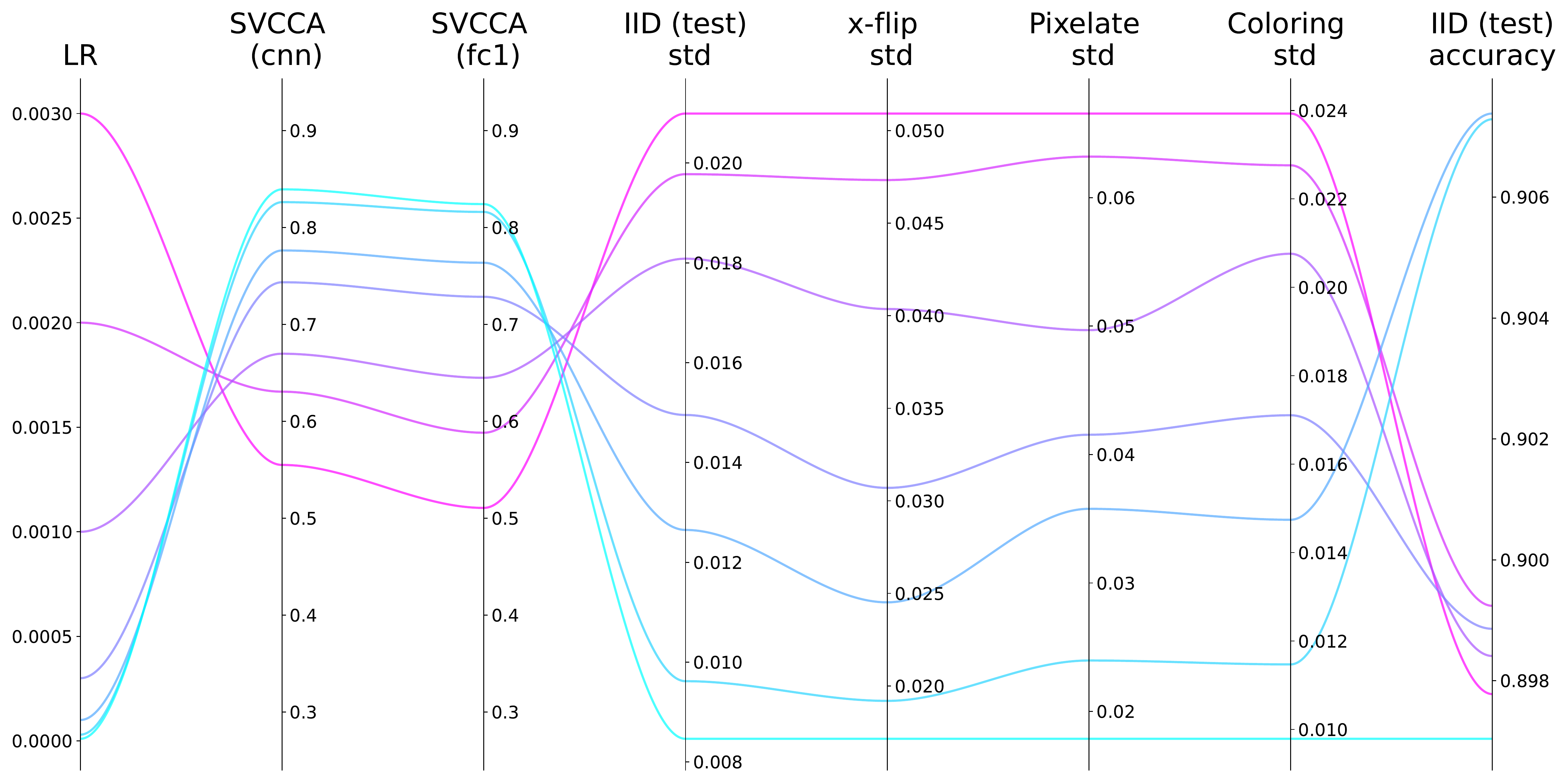}
\includegraphics[width=0.48\textwidth]{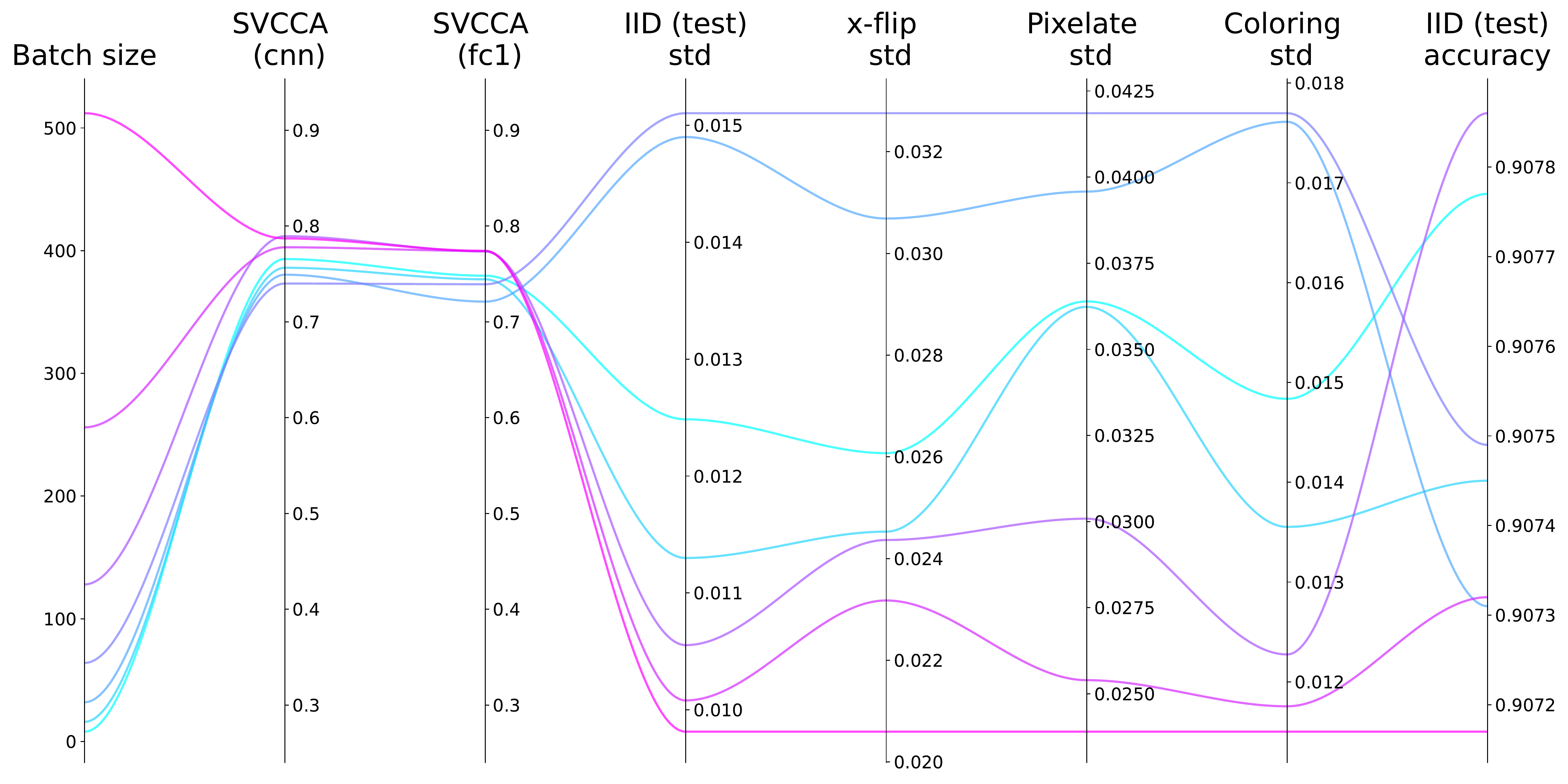}\\
\begin{tikzpicture}[inner sep=0]
    \node[rotate=90,minimum width=.99\figurewidth,inner sep=1pt,rounded corners=1pt] at (\figurewidth,.55\figurewidth) {\bf \scriptsize CIFAR-9};
\end{tikzpicture}
\includegraphics[width=0.48\textwidth]{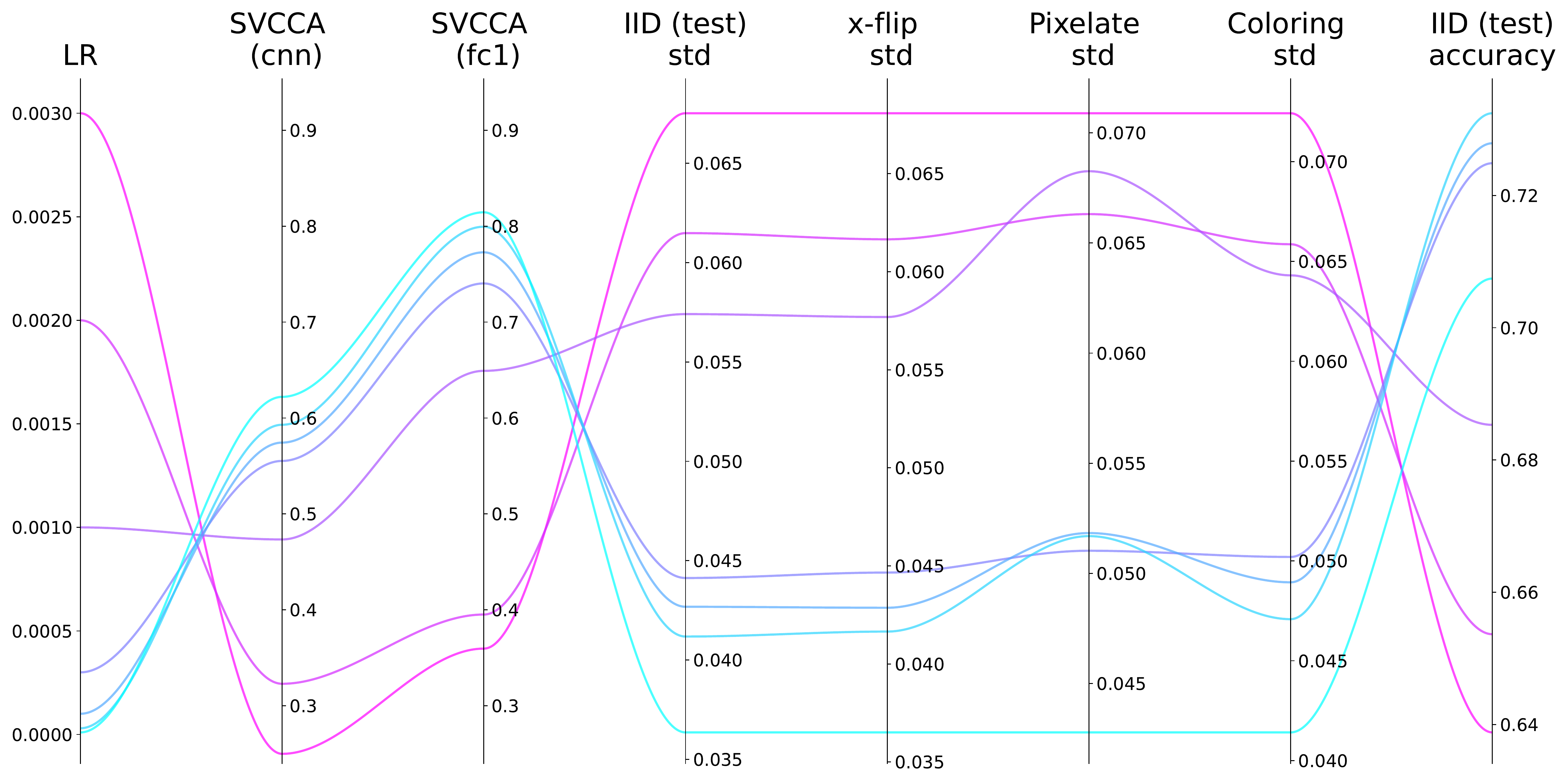}
\includegraphics[width=0.48\textwidth]{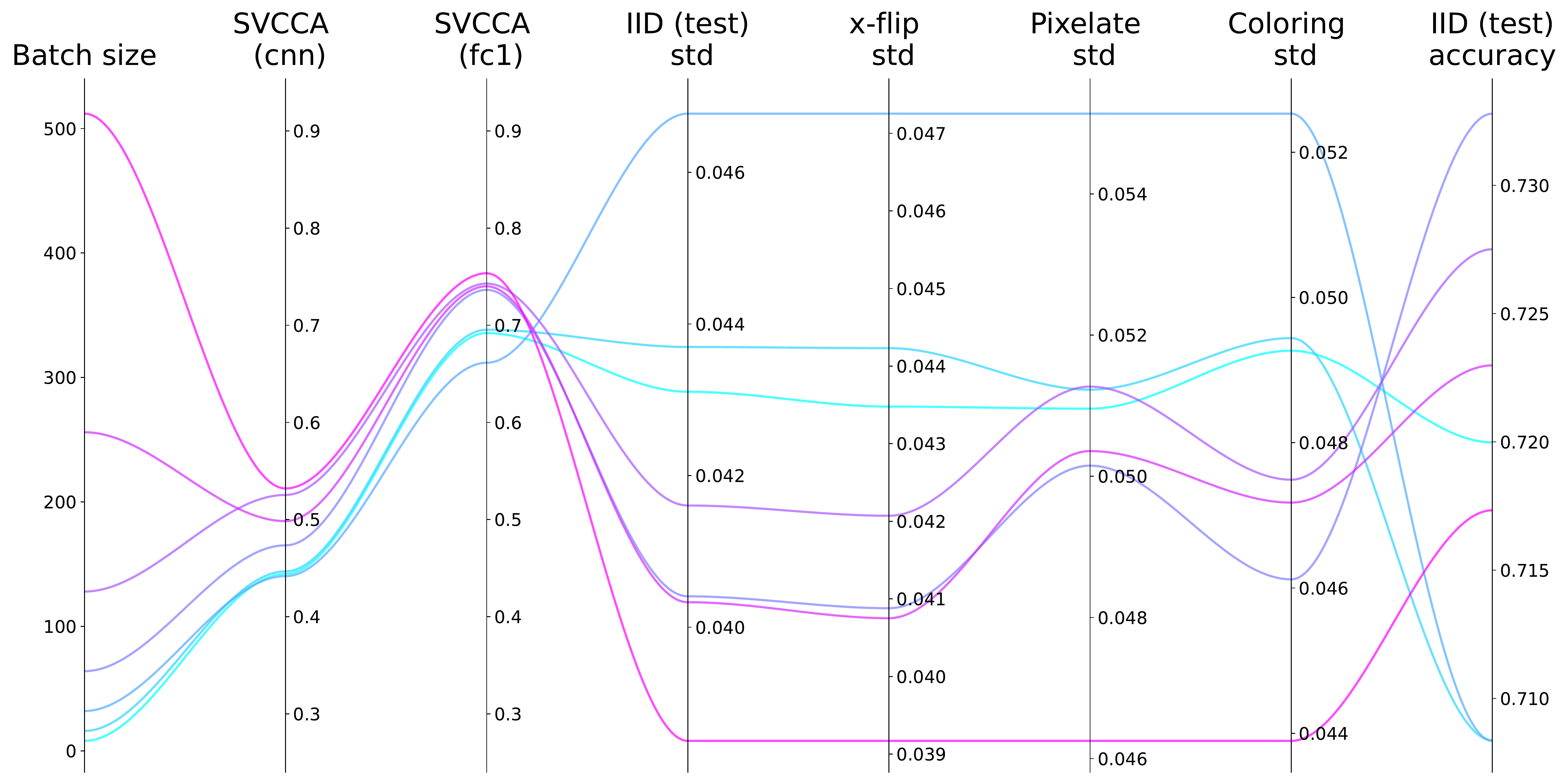}\\
\begin{tikzpicture}[inner sep=0]
    \node[rotate=90,minimum width=.99\figurewidth,inner sep=1pt,rounded corners=1pt] at (\figurewidth,.55\figurewidth) {\bf \scriptsize SVHN};
\end{tikzpicture}
  \includegraphics[width=0.48\textwidth]{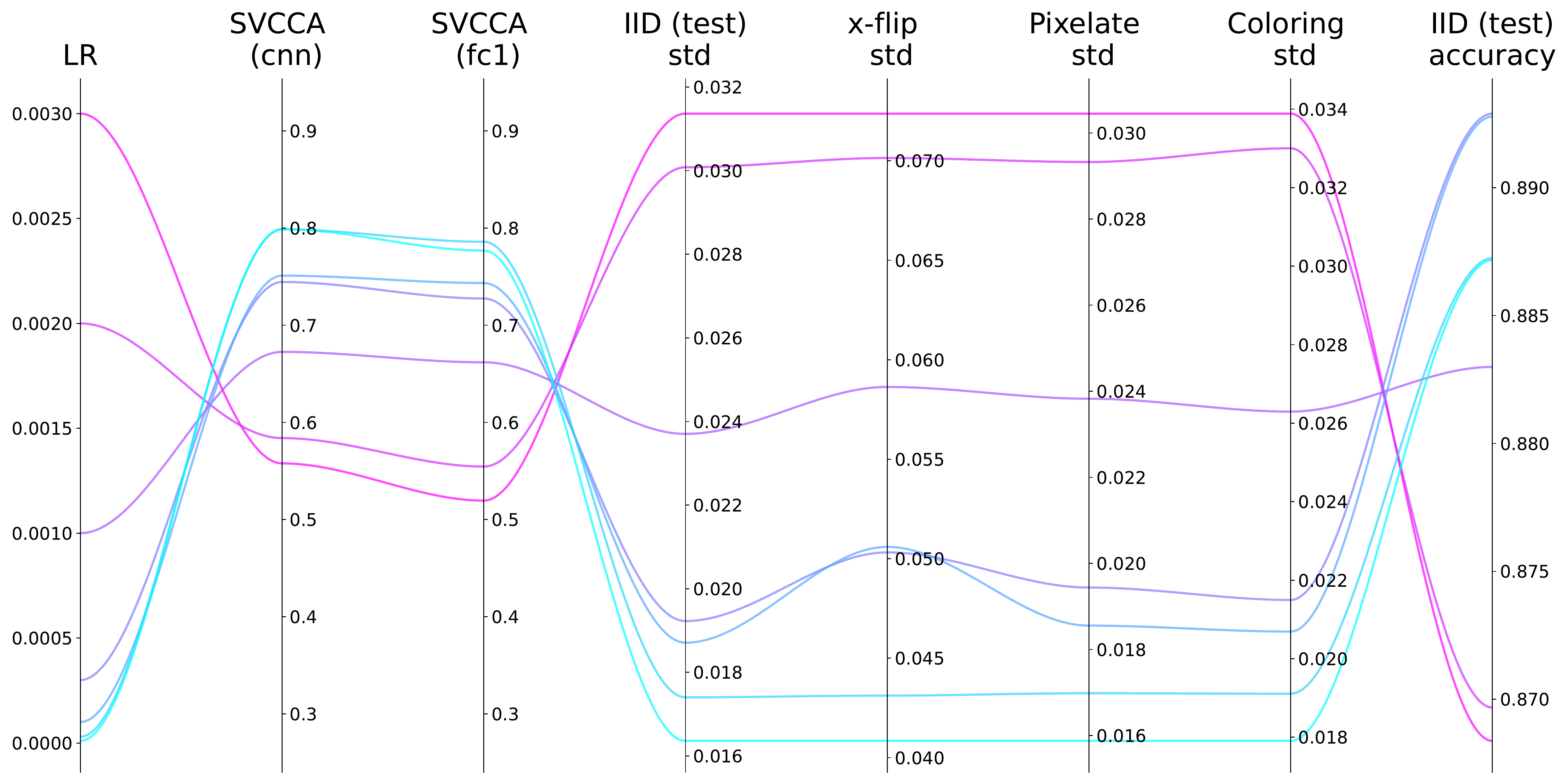}
  \includegraphics[width=0.48\textwidth]{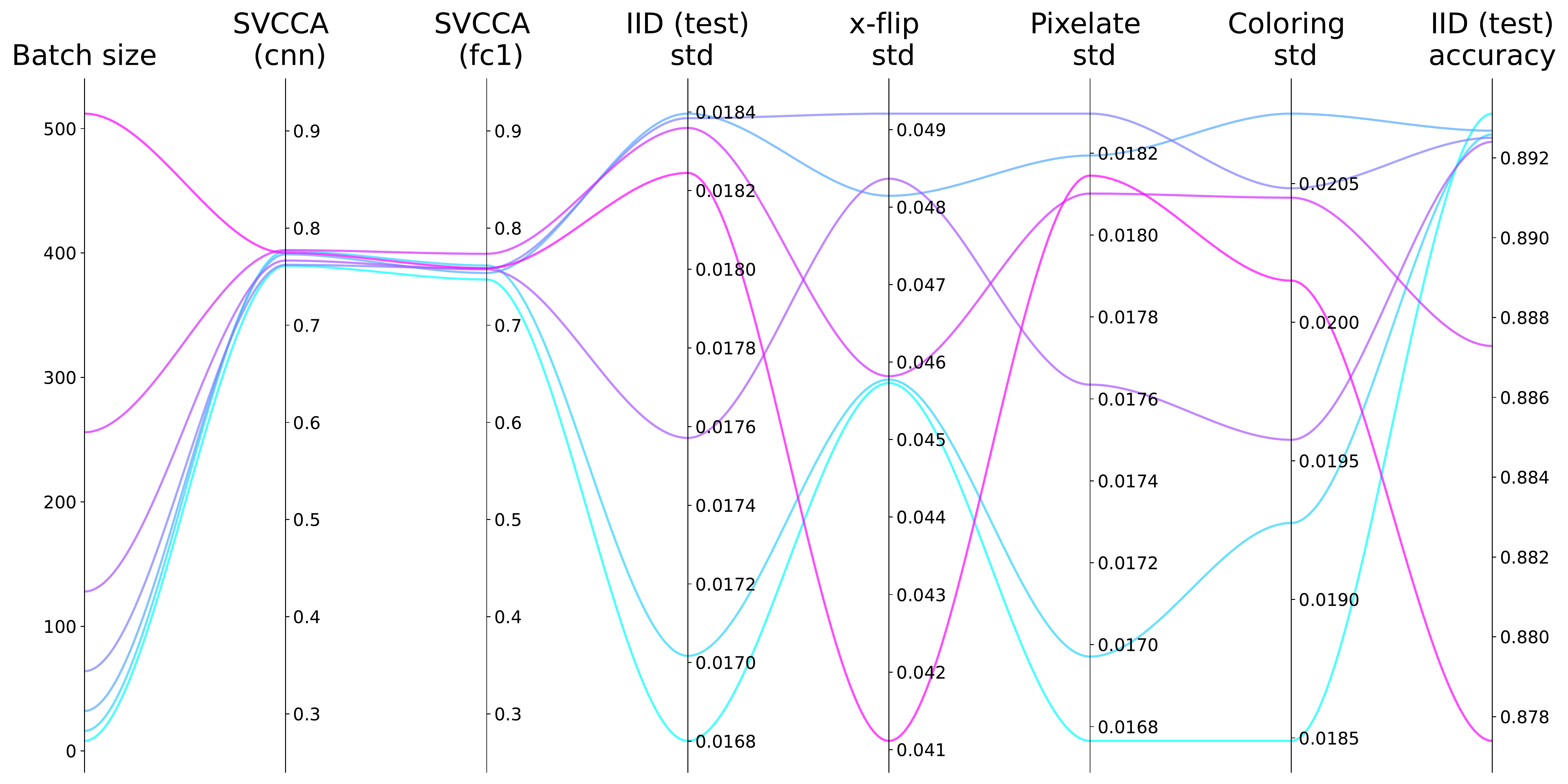}
 \begin{tikzpicture}[inner sep=0]
    \node[rotate=90,minimum width=.99\figurewidth,inner sep=1pt,rounded corners=1pt] at (\figurewidth,.55\figurewidth) {\bf \scriptsize MNIST};
\end{tikzpicture}
\includegraphics[width=0.48\textwidth]{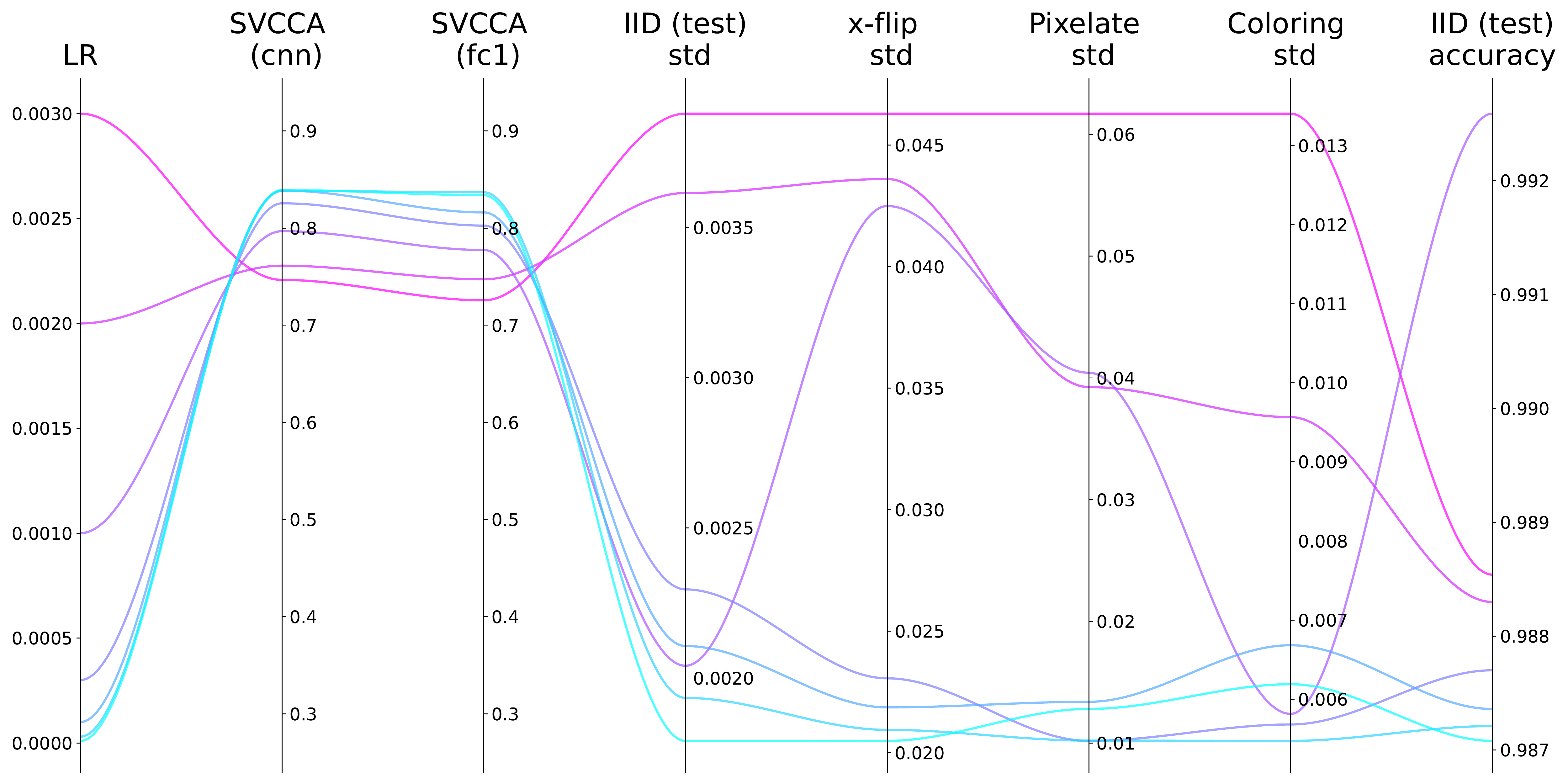}
\includegraphics[width=0.48\textwidth]{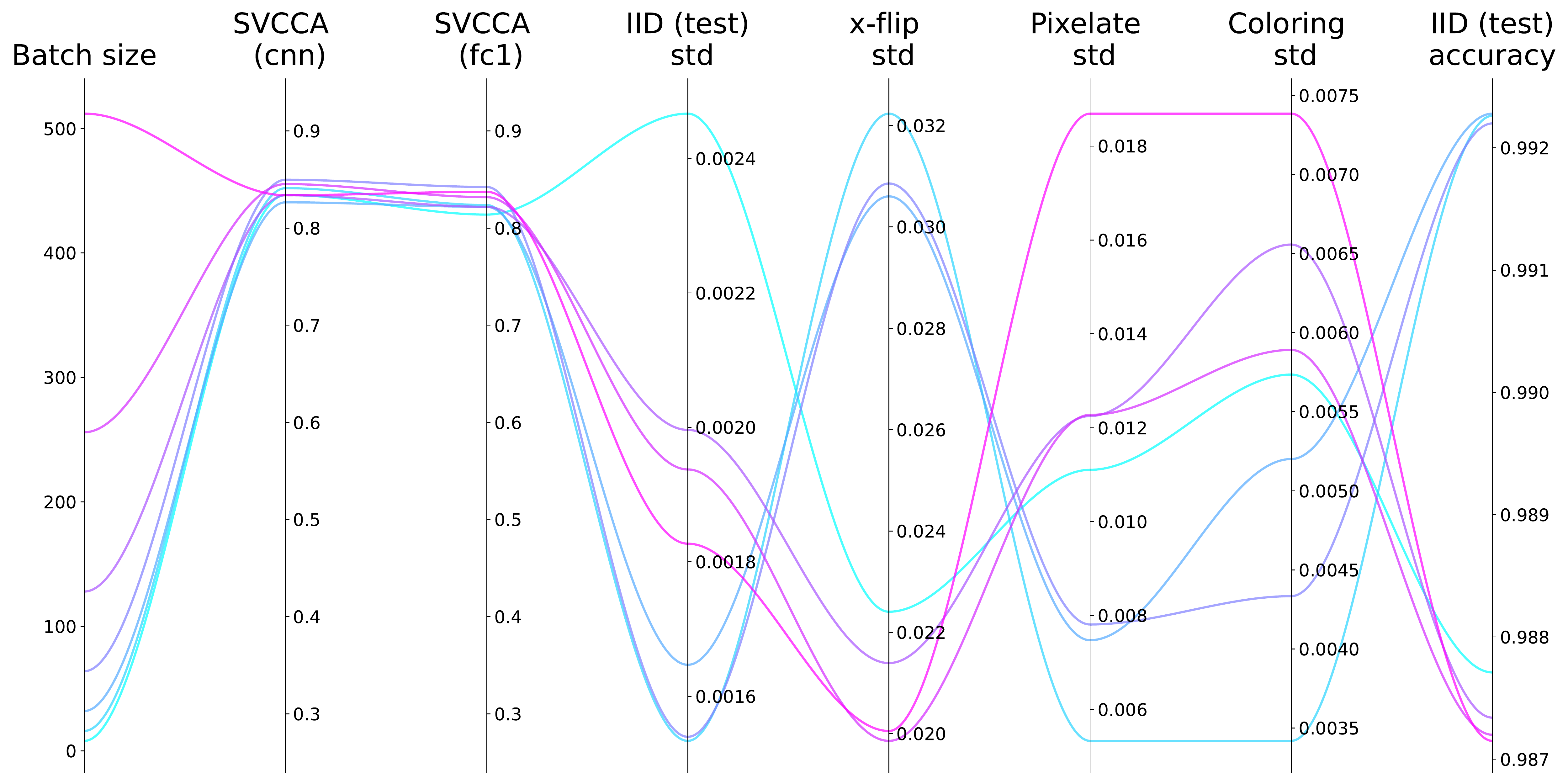}\\
  \caption{
  Results using the the pseudo-maximum test set accuracy.
  \label{fig:hpall}}
\label{fig:parallelbest}
\end{figure}

\end{document}